\documentclass[review]{elsarticle}
\usepackage{amsmath}
\usepackage{setspace}
\usepackage{ulem}
\usepackage[perpage]{footmisc}
\usepackage{lineno}
\usepackage{booktabs}
\usepackage{subcaption}

\usepackage{algorithm,tabularx}

\usepackage[noend]{algpseudocode}
\usepackage{tikz}
\usepackage{color}
\usepackage{csquotes}
\usepackage{dsfont}
\usepackage{xhfill}
\usepackage{lineno}
\usepackage{soul}
\usepackage{emptypage}
\usepackage{rotating}
\usepackage{multirow}
\usepackage{booktabs}
\usepackage{makecell}
\usepackage{setspace}
\usepackage{parselines}
\usepackage{lineno}
\usepackage{setspace}
\usepackage{hyperref}
\usepackage{bbding}

\biboptions{sort&compress}

\modulolinenumbers[5]

\newlength\tbspace
\newcolumntype{L}{l<{\hspace{\tbspace}}}

\makeatletter
\newcommand{\multiline}[1]{%
  \begin{tabularx}{\dimexpr\linewidth-\ALG@thistlm}[t]{@{}X@{}}
    #1
  \end{tabularx}
}
\makeatother
\algnewcommand{\do}{\textbf{do }}
\algnewcommand\Input{\item[\textbf{Input:}]}%
\algnewcommand\Output{\item[\textbf{Output:}]}%

\journal{Knowledge-Based Systems}
\PassOptionsToPackage{draft}{graphicx}

\begin{document}

\begin{frontmatter}

\title{ Multi-Expert Human Action Recognition with Hierarchical Super-Class Learning}

\author{Hojat Asgarian Dehkordi}
\ead{h\_asgariandehkordi@elec.iust.ac.ir}

\author{Ali Soltani Nezhad}
\ead{ali\_soltaninezhad@elec.iust.ac.ir}

\author{Hossein Kashiani}
\ead{hossein\_kashiyani@alumni.iust.ac.ir}

\author{Shahriar Baradaran Shokouhi}
\ead{bshokouhi@iust.ac.ir}
\author{Ahmad Ayatollahi}
\ead{ayatollahi@iust.ac.ir}


\address{School of Electrical Engineering, Iran University of Science and Technology, Tehran, Iran}

\begin{abstract}

In still image human action recognition, existing studies have mainly leveraged extra bounding box information along with class labels to mitigate the lack of temporal information in still images; however, preparing extra data with manual annotation is time-consuming and also prone to human errors. Moreover, the existing studies have not addressed action recognition with long-tailed distribution. In this paper, we propose a two-phase multi-expert classification method for human action recognition to cope with long-tailed distribution by means of super-class learning and without any extra information. To choose the best configuration for each super-class and characterize inter-class dependency between different action classes, we propose a novel Graph-Based Class Selection (GCS) algorithm. In the proposed approach, a coarse-grained phase selects the most relevant fine-grained experts. Then, the fine-grained experts encode the intricate details within each super-class so that the inter-class variation increases. Extensive experimental evaluations are conducted on various public human action recognition datasets, including Stanford40, Pascal VOC 2012 Action, BU101+, and IHAR datasets. The experimental results demonstrate that the proposed method yields promising improvements. To be more specific, in IHAR, Sanford40, Pascal VOC 2012 Action, and BU101+ benchmarks, the proposed approach outperforms the state-of-the-art studies by 8.92\%, 0.41\%, 0.66\%, and 2.11 \% with much less computational cost and without any auxiliary annotation information. Besides, it is proven that in addressing action recognition with long-tailed distribution, the proposed method outperforms its counterparts by a significant margin.


\end{abstract}

\end{frontmatter}
\section{Introduction}\label{sec.1}
Recently, human action recognition has attracted significant attention in computer vision due to its real-world applications. Action recognition in computer vision is generally categorized into action recognition in videos \cite{si2020skeleton,wu2021global,wang2021multi,ozyer2021human} and in still images \cite{dong2021knowledge,ji2019context,herath2019using}. Action recognition in still images aims to identify the type of human activity in a static image without any temporal information. Since a wide range of human activities such as \textsl{running} and \textsl{smoking} can be identified with a single input image and without extra motion cues, image-based action recognition has gained considerable attention. However, the lack of temporal information makes image-based action recognition more challenging rather video-based action recognition \cite{zheng2019spatial}. Human action recognition (HAR) has numerous applications such as sports analysis, abnormal behavior recognition, wildlife observation, image tagging, image retrieval, and human-computer interaction \cite{ozyer2021human,yadav2021review}. As still images lack temporal information, typically, action recognition studies extract spatial information from images. In this respect, initial studies in HAR have mostly employed low-level feature extraction techniques to capture low-level structures; however, they fail to achieve reliable and satisfactory results \cite{mohammadi2019ensembles}. Another category of approaches explores to leverage object detector \cite{kim2020detecting} or pose estimator \cite{mi2021pose} developments to detect the keypoint joints \cite{safaei2020ucf}, which are the most discriminative regions in the foreground area. Such detected areas favorably contribute to the overall action recognition accuracy. For instance, the detected areas in relation to the bicycle instance and the lower part of the human body could provide the methods with discriminative features in action recognition for \textsl{riding a bike} class.

In recent years, Convolutional Neural Networks (CNNs) have demonstrated their superior capabilities in several computer vision tasks and have emerged as a promising tool for HAR \cite{xiao2021federated,ozyer2021human,plizzari2021skeleton,yoshikawa2021metavd}. CNNs have significantly advanced the keypoint detection and estimation for action recognition. Nevertheless, the standard CNN-based studies mainly have three constraints. First, for the training phase, these studies need a large amount of annotations regarding the human bounding boxes or body parts as well as action labels. Second, though CNNs could extract rich feature hierarchies, they generally struggle to extract the structural features for modeling the relationship among human keypoints in action recognition. This is ascribed to the fact that all activities are treated uniformly without considering any correlation and similarity context. With this learning process, the misclassification between relevant activities such as \textsl{Reading a book} and \textsl{Writing a book} is penalized same as other irrelevant action activities such as \textsl{Fixing a bike}. Third, most studies in HAR conjecture a balanced class distribution in their training phase over the existing well-organized balanced datasets such as Stanford40 \cite{yao2011human}, Pascal VOC 2012 Action \cite{everingham2015pascal} and BU101+ \cite{ashrafi2021action} datasets. Nevertheless, human instances exist at various frequencies in different class activities naturally, and this makes the underlying class distribution of the real-world dataset severely imbalanced. Despite the fact that a large number of research activities have been carried out with respect to action recognition \cite{dong2021knowledge,liu2018loss,8214269,LI2020107341,mi2021pose}, to the best of our knowledge, none of them have taken into account the class imbalance issue in nature for action recognition in still images. This challenge impairs their performance when employed in large-scale real-world datasets. To address this challenge, we introduce a new large-scale Imbalanced Human Action Recognition dataset (IHAR) with a long-tailed distribution so that we can fairly assess the generalization of our proposed approach in comparison to the state-of-the-art studies.



In this paper, we propose a super-class learning approach for action recognition, named SCLAR, in still images. The merits of SCLAR are threefold: \textbf{1)} Our hierarchical SCLAR can detect visually distinct and subtle differences between various classes for action recognition. When it comes to the super-class learning, researchers attempt to promote inter-class variation between different classes. To reach this objective, we decompose the action recognition task into two-phase classification problem. First, a bucket of separate lightweight CNN classifiers (also called fine-grained classifiers) are pre-trained for the subsets of classes (i.e., super-classes). Then, a coarse-grained classifier is adopted to determine the relevant fine-grained classifiers in the first phase. The fine-grained classifiers ultimately would output the final classification label. With such a methodology, different activity classes are routed downstream to different fine-grained classifiers such that the inter-class variance among different classes increases. Thanks to the discriminative two-phase framework, specialized features are tuned to discriminate subtle visual differences in similar and different action classes from each other. \textbf{2)} Our action recognition framework surpasses the state-of-the-art approaches in HAR, while demanding much less computational cost and memory in that the coarse-grained and fine-grained classifiers in our framework are both compact models. As the action recognition task is a fine-grained classification, recent state-of-the-art studies require deeper CNN models with high capacity and computational complexity to detect subtle differences between similar classes. However, in a wide range of applications, we require low computation load for deployment on edge devices. SCLAR addresses this bottleneck observed in other studies while outperforming its competitors. \textbf{3)} Our framework can also address action recognition with long-tailed distribution. Since distinctive classes in super-class share knowledge in SCLAR, the under-represented classes manage to generalize better. Class imbalance occurs when there is considerable discrepancy among the cardinality of various classes \cite{suh2021cegan,kim2021novel}. In the wild, we mostly encounter an inherent imbalance issue concerning human activity classes in action recognition. This causes the training loss to be biased toward the over-represented classes while simultaneously rendering the under-represented classes unable to reflect intra-class variation. When the knowledge in the coarse-grained phase is transferred to the fine-grained phase through a hierarchical structure, different super-classes value specialized subsets of features in relation to the under-represented classes. Therefore, the learned representation can focus better on the specific classes in a super-class with low cardinality and hard samples. After all, our framework would ideally relieve the data imbalance issue in the action recognition task.

The rest of the paper is structured as follows: Section \ref{sec.2} provides a literature review of the most recent HAR research. Section \ref{sec.3} presents an elaborate description of the proposed two-phase multi-expert classification method. This section also provides a detailed explanation for the Graph-based Class Selection Method. Section \ref{sec.5} introduces our large-scale dataset for action recognition in still images with the long-tailed distribution. Section \ref{sec.6} offers exhaustive assessments of the proposed method on several datasets to gauge the contributions of different ingredients in our framework. Finally, section \ref{sec.7} concludes our research work.

\section{Related Works}\label{sec.2}

In this section, first, the studies in relation to CNN-based human action recognition are explained. Then, the methods equipped with an object or pose detector are addressed. Ultimately, the state-of-the-art studies regarding data imbalance issue are described in more detail.

\subsection{Still Image Action Recognition}

In recent years, numerous methods in visual object classification \cite{YANG2021103245}, object detection \cite{LI2021418}, object tracking \cite{wang2020robust}, and action recognition \cite{dong2021knowledge,ji2019context,wang2021multi} have been proposed based on CNNs. In image-based HAR, the state-of-the-art studies exploit CNNs along with object detection module to locate human agents and mitigate irrelevant noisy context. Authors in \cite{rosenfeld2018action} first detect different objects and their attributes and then compute a weighted sum of such detected instances for image-based HAR. Yan et al. \cite{YAN2017118} attempt to represent still images as a bag of image patches which are extracted by means of region proposal approaches. To do so, the FV encoding is applied to CNN features of image patches, and the spatial pyramid representation is employed for spatial feature extraction. In \cite{8214269}, the scene-level and region-level contexts are captured at the same time through a well-designed multi-branch attention networks. Two context branches for scene and region attention and a branch for target human region classification are incorporated into the proposed network for HAR.  The requirement of human bounding boxes in still images is relaxed for action recognition in \cite{ActionZhang}. For this relaxation, the detected object proposals via selective search are decomposed into fine-grained components, and the final action predictions are calculated using an efficient product quantization to take into account the human-object interaction areas.

\subsection{Object/Pose-Based Action Recognition}

To comprehend the visual world, we should detect individual object instances in a scene as well as their interactions. In this regard, studies in this category initially recognize the instances in still images and then employ the visual relationship between human and objects to detect different actions. Authors in \cite{gkioxari2018detecting} make use of an inferred target location to find the correct object concerning the specific action. To be more specific, they estimate an action-type specific density around the portion of target objects by means of a human-centric recognition branch in the Faster R-CNN model. Equipped with the keypoint detection network, Wang et al. \cite{wang2020learning} compute an interaction vector based on the human and object center points to directly determine interactions between human-object pairs. Ma et al. \cite{ma2020human} aim to promote features with a human-object relation module to calculate pair-wise interaction context between human and objects
for action classification.

As human bodies are structural objects, modeling human activities by means of motion context is feasible. The motion context concerning entire body structure or different body parts can provide valuable geometric interactions in various human activities, which are specifically applicable for fine-grained activity recognition. In this vein, authors in \cite{ZHAO2016134} recognize semantic regions in the human bounding box and arrange features of detected regions in a top-down spatial order to strengthen inter-class variance for higher discriminative representation. Mottaghi et al. \cite{MOTTAGHI2020921} propose a novel feature descriptor called Histogram of Graph Nodes (HGN), whereby the skeletons of the silhouettes can be converted into a graph to model the articulated human body skeleton. In \cite{zhao2017single}, human body is partitioned into seven parts such as a head with a few semantic part actions to classify human action category. For this objective, a CNN model with two subnetworks is utilized for part localization and action prediction. Li et al. \cite{li2020pastanet} propose a Hierarchical Activity Graph to encode human instances and their body part to reason out the activity classes using part-level semantics. A unified CNN model is adopted in \cite{LI2020107341} to capture structural details of body instances and integrate several body structure cues, namely Structural Body Parts and Limb
Angle Descriptor cues, for HAR.

While significant progress has been made in action recognition studies \cite{gkioxari2018detecting,wang2020learning,ma2020human,ZHAO2016134,MOTTAGHI2020921,zhao2017single,MOTTAGHI2020921,LI2020107341,li2020pastanet}, most of which require high computational resources beyond the capabilities of edge devices. However, our two-phase framework is constructed from lightweight CNNs, which is specifically tailored for resource-constrained platforms. As such, under severe constraints on computing power and memory resource, the proposed SCLAR is much faster than its counterparts.

\subsection{Data Imbalance Issue}

The imbalance class distribution would lead to poor generalization of CNNs, and the learning process with such a distribution would be biased toward over-represented classes. This eventually results in a biased CNN, which fails to detect the subtle discriminant features needed to classify the under-represented samples. Thus, the over-fitting in favor of the over-represented classes is inevitable for most approaches in action recognition. Though learning from imbalanced data has not been addressed in action recognition in still images, it has been well studied in other computer vision tasks such as object detection \cite{bria2020addressing,li2020overcoming} and image classification \cite{suh2021discriminative,suh2021cegan,kim2021novel}. To mitigate this deep-rooted issue, several conventional and also recent studies have been conducted, such as over-sampling the under-represented samples and under-sampling the under-represented samples. Also, recent studies include Prime Sample Attention \cite{cao2020prime}, AP Loss  \cite{chen2019towards}, DR Loss \cite{qian2020dr}, pRoI Generator \cite{oksuz2020generating}, and IoU-based Sampling \cite{pang2019libra}. While other studies in HAR have not considered the imbalanced data issue, in this paper, different classes are selected and categorized into super-classes to handle the data imbalance issue. With this perspective, the knowledge of different classes would be shared during the training procedure in order that the overall gradient in the training phase would not be dominated by the over-represented classes.

\begin{figure}[!t]
  \centering
  \includegraphics[scale=0.65]{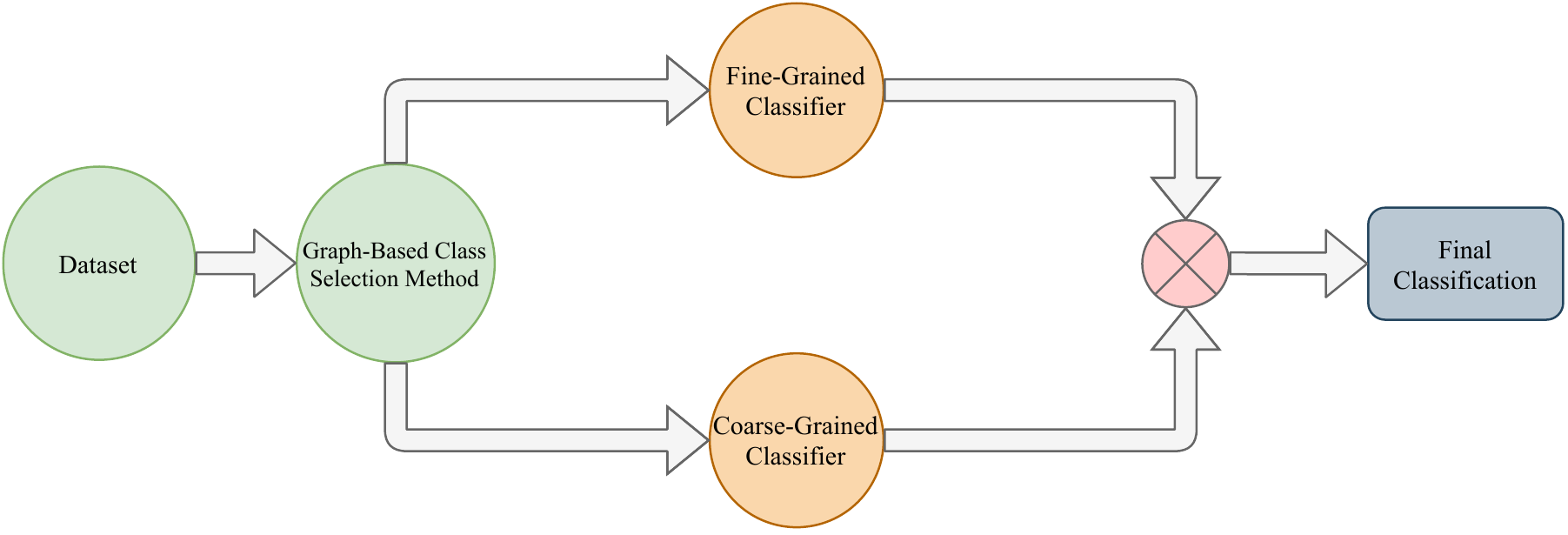}
  \caption{Flowchart of the proposed hierarchical action recognition.}\label{fig.flow}
\end{figure}

\section{The Proposed Method}\label{sec.3}
Recall that in this work, a bucket of separate lightweight CNN classifiers is required to be pre-trained for different super-classes. To this end, the entire dataset is first divided into \textsl{M} super-classes in such a way that all super-classes contain a relatively balanced distribution. Thanks to this dataset division, the previously condensed inter-class boundary would be smoothed. The EfficientNet-B0 model \cite{tan2019efficientnet} is adopted as the lightweight CNN classifier in the fine-grained classification phase. The EfficientNet models are all pre-trained by means of input data in various classes of their relevant super-class. As such, they would become specialized in capturing domain-specific features in relation to the classes of their corresponding super-classes. A combination of these lightweight expert backbones constructs the fine-grained stage of our architecture configuration for the final action recognition prediction. Finally, the most relevant backbones for Fine-Grained Classifier (FGC) are activated by means of the Coarse-Grained Classifier (CGC). It should be noted that the performance of EfficientNet models would drop with a small number of super-classes (i.e., \textsl{M}). On the other side, a large number of super-classes would incur computation overhead. Thus, to achieve the optimal trade-off between classification performance and computational cost, $M = {3,4,5} $ super-classes are ablated in our assessments. Moreover, it should be stressed that the configuration of each super-class is one of the major factors in the performance of our two-phase architecture. Therefore, the GCS algorithm is proposed in this study to probe the best configuration for each super-class whereby the input embedding space is partitioned into several super-classes, and the inter-class variation would be increased. Figure \ref{fig.flow} illustrates the flowchart of the hierarchical action recognitor in our method. First, the GCS algorithm pinpoints the best configuration of each super-class. Then, CGC selects the best pre-trained backbone expert and activates FGC. Finally, FGC determines human action recognition classes.

\subsection{Fine-grained Phase}

\begin{figure*}[!t]
\captionsetup[subfigure]{justification=centering}
    \centering
    \begin{subfigure}[b]{\textwidth}
        \centering
        \includegraphics[width=.9\linewidth]{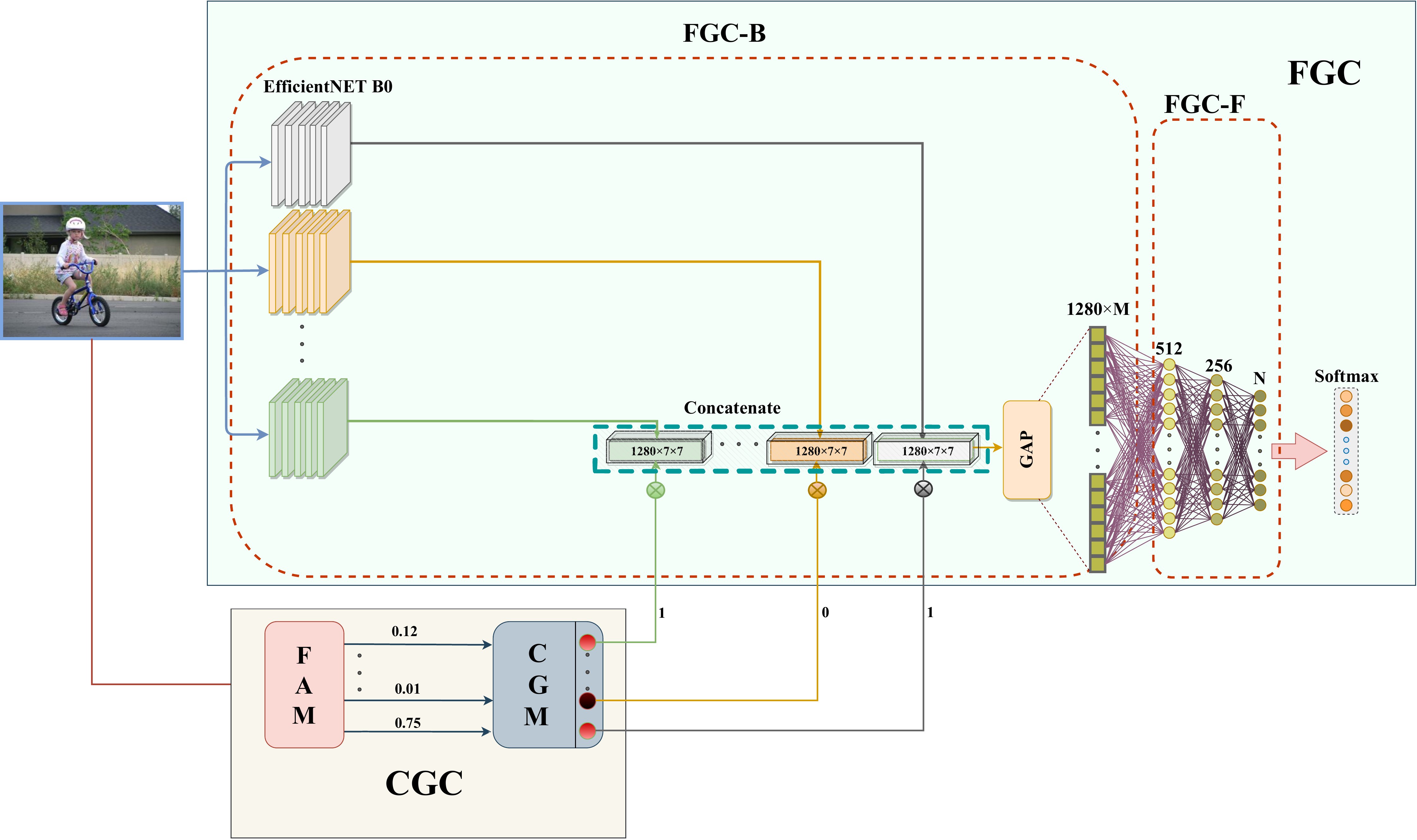}%
        \vspace{.5mm}
        \caption{}
    \end{subfigure}
    \vskip\baselineskip
    \begin{subfigure}[b]{\textwidth}
        \centering
        \includegraphics[width=.8\linewidth]{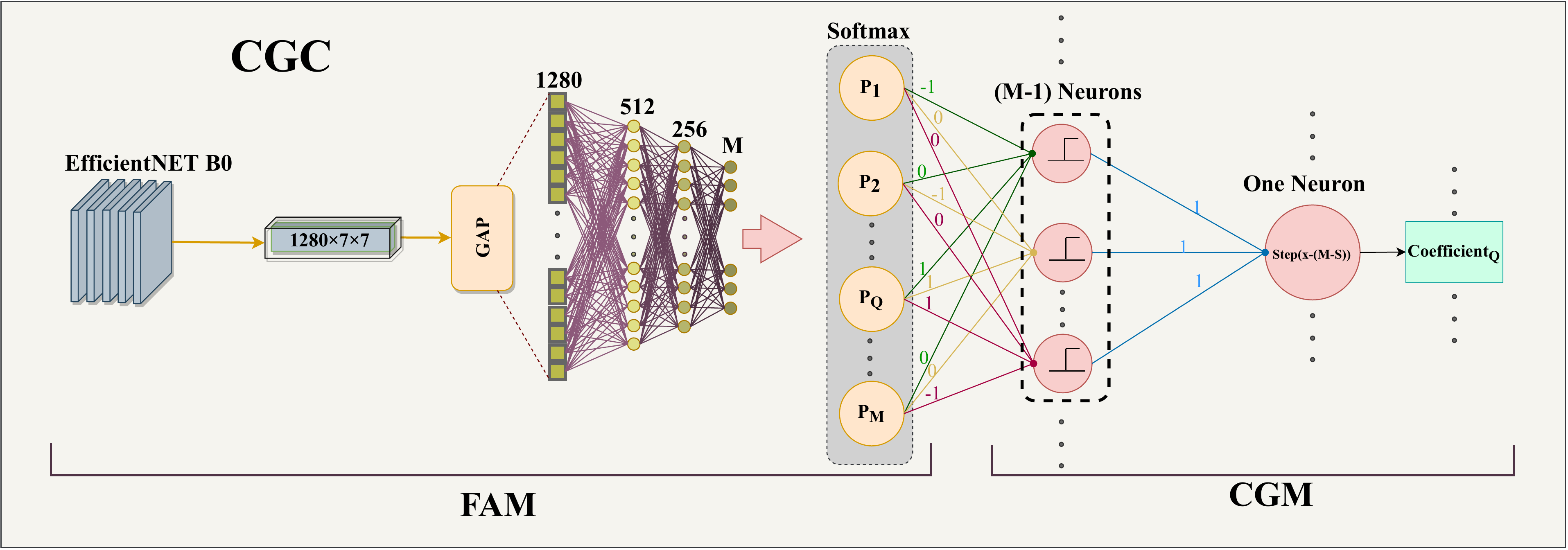}%
        \caption{}
    \end{subfigure}
  \caption{The proposed architecture of our two-phase hierarchy. (a) Network architecture. (b) The coarse-grained classifier. }
\label{fig.architecture}
\end{figure*}

Figure \ref{fig.architecture} (a) depicts the entire architecture of FGC in the fine-grained classification phase. The FGC is composed of two components, namely FGC-B and FGC-F (B and F denote backbone and fully connected layer components). The FGC-B is constructed from $M$ EfficientNet-B0 models, which can readily scale up conventional CNNs in a more principled manner to any resource limitations with a desired efficiency. For more details about the architecture of EfficientNet-B0, such as kernel size and channel size in convolution filters, we refer the readers to \cite{tan2019efficientnet}. In the fine-grained classification phase, first, all EfficientNets are applied to the input image and each of which outputs a $7 \times 7 \times 1280$ feature map. The specialized output maps corresponding to different EfficientNets are then concatenated channel-wise and fed into the CGC to be elected for the remaining section of the FGC. Then, the compact subnetwork (FGC-F) runs on the specialized feature map. The global averaging pooling (GAP) is also employed after the concatenation step to regularize the FGC and mitigate the over-fitting issue. Lastly, to train the FGC-F, the softmax loss is utilized.


\subsection{Coarse-grained Phase}
As represented in Figure \ref{fig.architecture} (a), to learn discriminative features for each super-class, CGC determines the pre-trained expert backbones for the relevant FGC-F in the fine-grained classification phase. The CGC empowers FGC to differentiate between subtle variations between different action classes. That is to say, it would help the FGC to strengthen inter-class discrepancies. This is due to the fact that the pre-trained backbones in the FGC-B have been separately tailored for different classes. For selecting the expert backbones, the output probabilities are produced by the Feature Attention Module (FAM), which is made up of EfficientNet-B0 and has been previously trained with softmax loss. The output probabilities of different experts over the corresponding super-classes are then converted to Top-S scores and S-hot masks that are point-wisely multiplied by the FGC-B. We opt for the Top-2 score rather than the Top-1 score since it hedges more relevant fine-grained backbones into a strong one and aggregates the most relevant subset-specific features. Also, the ablation study verifies the effectiveness of Top-2 score. The Coefficient Generation Module (CGM) addresses this task by assigning ones and zeros to the relevant and irrelevant fine-grained backbones in the FGC-B, respectively. To demonstrate this procedure, the FAM is applied to $M$ super-classes. This results in the output probabilities $P = \{P_{1}, P_{2},...,P_{M}\} $, where $P_{Q}$ corresponds to the probability of super-class $Q$ ($1 \leq Q\leq M$) . Given that FAM generates the probabilities for the selection process in the FGC-B and $M$ denotes the number of all super-classes, the CGM would set output $Q$ to one when $P_{Q}$ is greater than at least $M-S$ number of output probabilities. The number of probabilities less than $P_{Q}$ can be computed by step function as follows:

\begin{equation}\label{eq.1}
\mathrm{u_{Q}}=\sum_{v=1 \atop v \neq Q}^{M} \operatorname{step}\left(P_{Q}-P_{v}\right),
\end{equation}

\noindent where $u_{Q}$ denotes the number of probabilities which are surpassed by $P_{Q}$. This equation compares the probability of $P_{Q}$ with other output probabilities $P_{v}$ and accumulates all the outputs of step functions. Then, the resulting coefficient corresponding to $P_{Q}$ (i.e., $c_{Q}$) would be set to one when $u_{Q}$ exceeds $M-S$ as follows:


\begin{equation}\label{eq.2}
\operatorname{c_{Q}} =\operatorname{step}\left(u_{Q}-(M-S)\right)= \begin{cases}1 & \mathrm{M}-\mathrm{S}<u_{Q} \\ 0 & \mathrm{M}-\mathrm{S} \geq u_{Q}\end{cases}
\end{equation}

\noindent To formulate the general form of Equation \ref{eq.1}, a weight $(M-1)\times 1 $ matrix is defined as follows:

\begin{equation}\label{eq.3}
\operatorname{step}\left(P_{Q}-P_{v}\right)=\left[\begin{array}{c}
\operatorname{step}\left(P_{Q}-P_{1}\right) \\
\operatorname{step}\left(P_{Q}-P_{2}\right) \\
\vdots \\
\vdots \\
\operatorname{step}\left(P_{Q}-P_{M}\right)
\end{array}\right]_{(M-1) \times 1},
\end{equation}

\noindent where each element draws a comparison between $P_{Q}$ and other output probabilities.

\pagebreak
Equation \ref{eq.3} can be also reformulated as follows:
\begin{equation}\label{eq.4}
\begin{gathered}
\operatorname{step}\left(P_{Q}-P_{v}\right)\\
=\operatorname{step}\left(\left[\begin{array}{ccccccc}
-1 & 0 & \ldots & 1 & 0 & \ldots & 0 \\
0 & -1 & \ldots & 1 & 0 & \ldots & 0 \\
\vdots & \vdots & \ddots & \vdots & \vdots & \vdots & \vdots \\
0 & \ldots & -1 & 1 & 0 & \ldots & 0 \\
\vdots & \vdots & \vdots & \ddots & \vdots & \vdots & \vdots \\
0 & \ldots & 0 & 1 & 0 & \ldots & -1
\end{array}\right]_{(M-1) \times(M)} \times\left[\begin{array}{c}
P_{1} \\
P_{2} \\
\vdots \\
P_{Q} \\
\vdots \\
P_{M}
\end{array}\right]_{(M) \times 1}\right)\\[10pt]
=\operatorname{step}\left(W_{1}^{T} \times P\right),
\end{gathered}
\end{equation}

\noindent where $W_{1}^{T}$ is a constant matrix which is multiplied by $P$ and ultimately results in $P_{Q}-P_{v}$. To implement Equation \ref{eq.4}, we use an MLP layer with $(M-1)$ neurons. The step function is adopted for the activation function of each neuron. Then, the weights of different neurons can be expressed as $W_{1}^{T}$, in which each row corresponds to a single neuron. Recall that the input of Equation \ref{eq.3} is the probabilities that have been produced by FAM. After comparing $P_{Q}$ with all other output probabilities and applying the step function to them, the computed output can be assessed in comparison with $M-S$ in Equation \ref{eq.2}. Thus, by unifying Equations \ref{eq.2} and \ref{eq.4}, we can finally obtain the final discrete coefficients as follows:
\pagebreak

\begin{equation}\label{eq.5}
\begin{gathered}
\operatorname{c_{Q}}= \operatorname{step}\left(\sum_{v=1 \atop v \neq Q}^{M} \operatorname{step}\left(P_{Q}-P_{v}\right)-(M-S)\right) \\[10pt]
\begin{array}{c}
=\operatorname{step}\left (\left[\begin{array}{llll}
1 & \ldots &  1
\end{array}\right]_{1 \times(M-1)} \times\left[\begin{array}{c}
\operatorname{step}\left(P_{Q}-P_{1}\right) \\
\operatorname{step}\left(P_{Q}-P_{2}\right) \\
\vdots \\
\vdots \\
\operatorname{step}\left(P_{Q}-P_{M} \right)
\end{array}\right]_{(M-1) \times 1 }-(M-S)\right)
\end{array}\\[20pt]
=\operatorname{step}\left(\left(W_{2}^{T} \times \operatorname{step}\left(P_{Q}-P_{v} \right)\right)-(M-S)\right),
\end{gathered}
\end{equation}

\noindent where $c_{Q}$ indicates the final coefficient for $P_{Q}$, and $W_{2}^{T}$ denotes an all-ones matrix. Same as Equation \ref{eq.4}, we can implement Equation \ref{eq.5} by means of single-layer MLP with one neuron and $\operatorname{step}\left(U-(M-S)\right)$ activation function. Figure \ref{fig.architecture} (b) illustrates the two-layer MLP network that performs the prementioned operations in CGM and outputs the discrete coefficients to select the best features in the FGC-B. Note that the two-layer MLP network is reused for individual M outputs.

\subsection{Graph-based Class Selection Method}\label{sec.4}

\begin{figure}[!t]
  \centering
  \includegraphics[scale=.6]{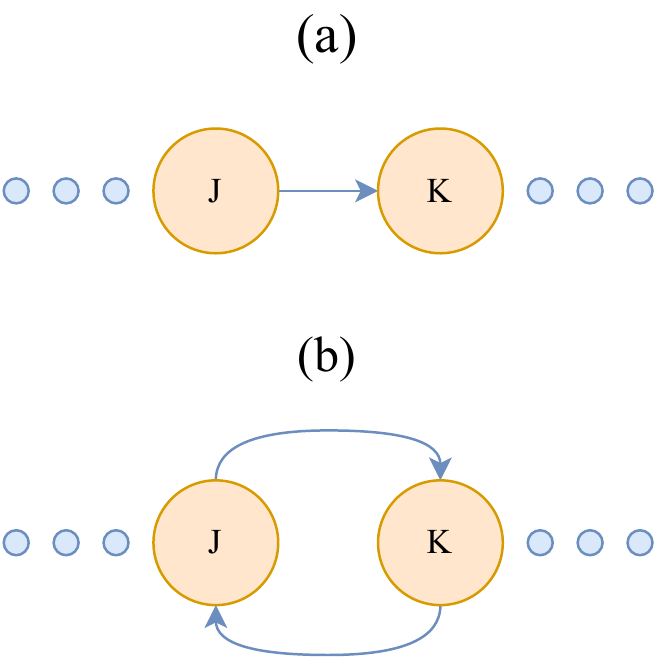}
  \caption{Two types of connections between different nodes in GCS algorithm. (a) one-way connection. (b) two-way connection.}\label{fig.GCM_connections}
\end{figure}

In action recognition in still images, there exist compact inter-classes boundaries among various activity classes. The proposed GCS method aims to partition the input dataset into M super-classes with relatively balanced distribution so that the inter-class variance would be strengthened desirably. To obtain the optimal configuration for each super-class in the GCS approach, $N$ classes are divided into M super-classes  according to their dependencies. To evaluate the inter-class dependency in a dataset and categorize correlated classes into distinct super-classes, a baseline network (coined BN) similar to the fine-grained expert classifiers (EffitientNets) is first pre-trained over the input dataset. Then, we can calculate the required dependencies between various classes in our dataset using the BN. More specifically, the pre-trained BN is first applied to all training images in class $j$; then, the output scores are sorted in ascending order. It is evident that the generated Top-1 scores correspond to class $j$. Then, the remaining $i$-th top scores for the majority of images in class $j$ would be related to class $c_i$, where $ 1\leq c_i \leq N $. The resulting $i$-th top scores ($i>1$) would determine the desired dependencies, which are required for partitioning different classes in the GCS algorithm. The second-order, third-order, and fourth-order dependencies for each class can be expressed as follows:

\begin{equation} \label{eq.6}
D_{2}=\left\{d_{21}, d_{22}, d_{23}, \ldots . . d_{2 N}\right\},
\end{equation}
\begin{equation} \label{eq.7}
D_{3}=\left\{d_{31}, d_{32}, d_{33}, \ldots . . d_{3 N}\right\},
\end{equation}
\begin{equation} \label{eq.8}
D_{4}=\left\{d_{41}, d_{42}, d_{43}, \ldots . . d_{4 N}\right\},
\end{equation}

\noindent where $D_{i}$ denotes $i$-th dependency set in the input dataset, and $d_{ij}$ indicates the $i$-th top score ($i>1$) corresponding to class $j$. Algorithm \ref{alg} summarizes the pseudocode for $D_{i}$ computation. For each dependency set $D_{i \in \{{2,3,4}\}}$ in the proposed GCS algorithm, the classes are represented as different nodes in a graph configuration, wherein the connections are based on the class dependencies and correlations. To model three dependency sets $D_{i \in \{{2,3,4}\}}$ in the GCS algorithm, two types of directed connections, namely one-way or two-way connections, are employed. In the GCS configuration, the second-order dependency is modeled either with the one-way or two-way connection (as depicted in Figure \ref{fig.GCM_connections}). It is worth noting that the third-order and fourth-order dependencies can only be represented by the one-way connection (Figure \ref{fig.GCM_connections}). Once class $j$ is dependant on class $k$, similarities between classes $j$ and $k$ can be formulated into four categories contingent upon the connection configuration, i.e., the order of dependencies, as below:

\begin{spacing}{1}
 \linespread{1.25}
\begin{algorithm}[tb]
  \begin{algorithmic}[1]
  \setstretch{1.35}
  \scriptsize
    \Input{Images, EfficientNet, Number of Classes $N$.}
    \Output{Dependencies order $D_{o}$, $o \in \{2,3,4\}$}
     \State Pre-train EfficientNet with Input Images.
    \State $D_{o} = []$
    \For {$i = 1, 2, ...N$  }
        \State \multiline{ $C_j =[]$, $j \in \{2,3,4\}$}
    \State $ClassImgs = Images[i]$.
       \For {img in ClassImgs}
       \State score = EfficientNet(img)
        \State $score = sort(score)$
        \State $C_r.add(index(score[r]))$, $r \in \{2,3,4\}$
    \EndFor
    \State $d_{m,i} = MaxIter(C_m)$, $m \in \{2,3,4\}$
    \State $D_n.add(d_{n,i})$, $n \in \{2,3,4\}$
    \EndFor\hrulefill
    \setcounter{ALG@line}{0}
    \State \textbf{Function}  MaxIter(A):
\State\hspace{0.5cm}{$F = zeros(length(A))$}
\State\hspace{0.5cm} \textbf{for}  {k  in range (1, length(A))}
\State \hspace{1cm}$F[A[k]] = F[A[k]] + 1$
\State\hspace{0.5cm}$F = sort(F)$
    \State \textbf{Return} $index(F[1])$
  \end{algorithmic}
  \caption{GCS Algorithm}
  \label{alg}
\end{algorithm}
\end{spacing}

\begin{equation} \label{eq.9}
{d_{2j}} = k ,\; {d_{2k}} = j \Rightarrow Type-1\; \; similarity \;\Rightarrow S_{1},
\end{equation}

\begin{equation} \label{eq.10}
{d_{2j}} = k , \;{d_{2k}}\neq j  \Rightarrow Type-2\; \; similarity\; \Rightarrow S_{2},
\end{equation}

\begin{equation} \label{eq.11}
{d_{3j}} = k , \;{d_{3k}}\neq j  \Rightarrow Type-3\; \; similarity\;\Rightarrow S_{3},
\end{equation}

\begin{equation} \label{eq.12}
{d_{4j}} = k ,\; {d_{4k}}\neq j \Rightarrow Type-4\; \; similarity\; \Rightarrow S_{4},
\end{equation}

\noindent where $d_{ij}$ represents the $i$-th top score corresponding to class $j$, which is dependant on class $k$. $ S_{i}$ is the status of similarities among different nodes. Clearly, $S_{t \in \{{1,2,3,4}\}}$ influences the performance of BN with different scales. To gauge the impact of $S_{t}$ on the final classification error, first, all class pairs with similarity $S_{t}$ are extracted for the input dataset. Then, to  calculate the classification error, we separately train a new EffitientNet for the selected class pairs with $S_{t}$. Finally, the average error for $S_{t}$ would be computed. These operations are formulated as follows:

\begin{equation} \label{eq.13}
E_{t} = \frac{1}{   \mid  N_{t} \mid}\sum_{\ell=1}^{\mid N_{t} \mid}{E_{\ell}},
\end{equation}

\begin{equation} \label{eq.14}
{E_{\ell}} = \frac{1}{\mid  K_{\ell} \mid+ \mid  J_{\ell} \mid} \sum_{\gamma=1}^{\mid  K_{\ell} \mid + \mid  J _{\ell}\mid} {y_{\gamma} \times \log({\hat{y}}_{\gamma})},
\end{equation}


\noindent where $E_{t}$ is the average error for newly trained backbones on the selected class pairs with $S_{t \in \{{1,2,3,4}\}}$. $N_{t}$ denotes the number of selected class pairs at each similarity level $S_{t}$. ${\hat{y}}_{\gamma}$ and $y_{\gamma}$ indicate ground-truth labels and the predicted label generated by the newly trained backbones. $\mid  K_{\ell}\mid$ and $ \mid  J_{\ell} \mid$ correspond to the number of images in the selected pairs $\ell$. Now, using Equations \ref{eq.13} and \ref{eq.14}, we can compute the average error $E_{t}$ for all similarities. The following inequality summarizes the obtained results in our evaluations as:

\begin{equation} \label{eq.15}
E_{1} > E_{2} > E_{3}>E_{4} > E_{0},
\end{equation}

\noindent where $E_{0}$ denotes the average classification error for the class pairs without any similarities. To partition the input space such that similar classes with negligible inter-class variation would be placed into different super-classes, the assigning process is carried out under the guidance of Equation \ref{eq.15}. Thus, the number of class pairs with $S_{1}$, $S_{2}$, $S_{3}$, and $S_{4}$ similarities would be minimized in the super-classes, respectively. For the partitioning phase in the GCS algorithm, since $S_{1}$, and $S_{2}$ are concerned with $D_{2}$ graphs, different classes in $D_{2}$ are first assigned to $M$ super-classes based on the structure of their connections. In this assignment stage, the number of $S_{1}$ and $S_{2}$ in each sub-class would be minimized with the first and second priorities. Afterward, concerning the neighboring nodes in the $D_{3}$ graphs, the configuration of subclasses are reformed to minimize the number of $S_{3}$ without constructing new $S_{1}$, and $ S_{2}$. In the end, the same reconfiguration is conducted for the neighboring nodes in $D_{4}$ graphs to minimize the number of $S_{4}$ without constructing new $S_{1}$, $ S_{2}$, and $S_{3}$. For a better understanding of the GCS algorithm, we provide an example in \ref{sec.app}.

\section{IHAR dataset}\label{sec.5}
\begin{figure}[!t]
  \centering
  \includegraphics[scale=.37]{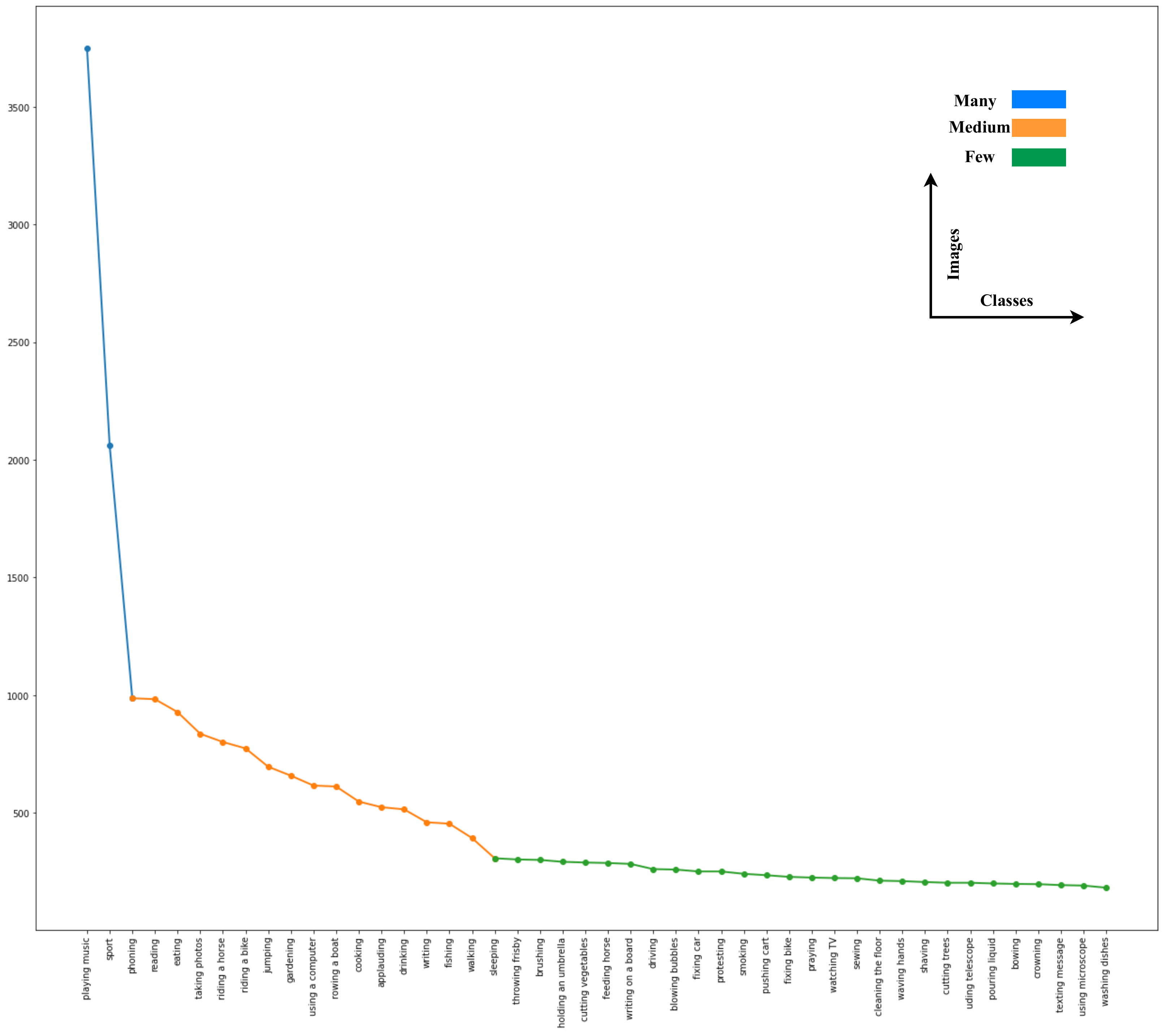}
  \caption{Distribution of different classes in the introduced IHAR dataset.}\label{fig.distribution}
\end{figure}

\begin{figure}[!t]
  \centering
  \includegraphics[scale=.27]{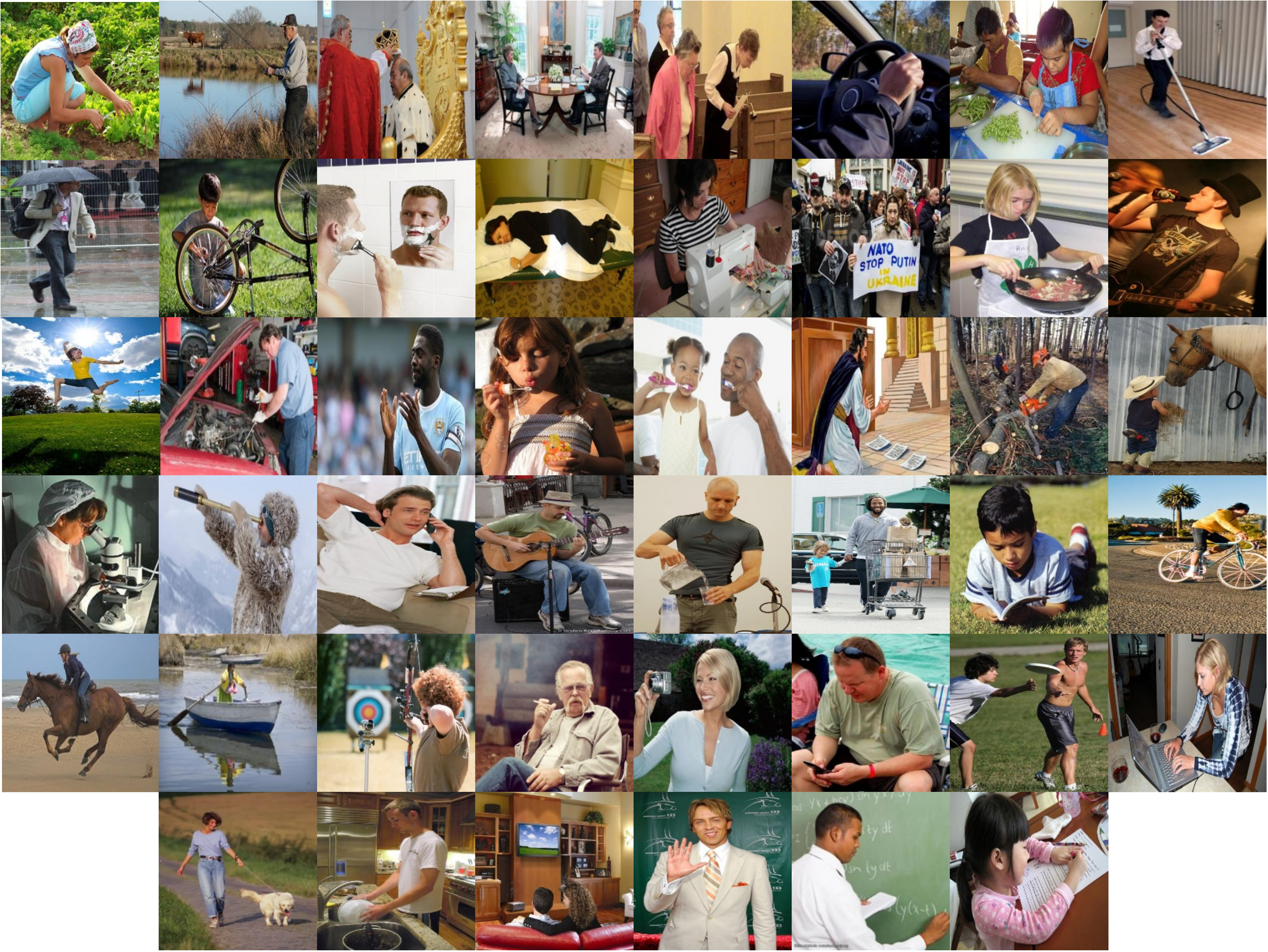}
  \caption{Different action classes in the introduced IHAR dataset.}\label{fig.IHARD}
\end{figure}

Real-world data often contain a long-tailed and open-ended distribution, and the data required for action recognition systems is no exception. This is due to the fact that human instances are present at varying rates in nature. An action recognition system requires to classify various action types among under-represented and over-represented classes and also generalize from a few known instances in under-represented classes. Previous well-known datasets in this field have not paid much attention to this requirement and consequently could easily struggle to keep working in long-tailed distribution. In this research, we introduce an imbalanced action recognition dataset with long-tailed distribution to fairly demonstrate the performance of our model in long-tailed distribution. Figure \ref{fig.distribution} illustrates the long-tailed distribution of the IHAR dataset. The IHAR dataset consists of 23854 images for 46 human action classes with a different number of samples. Different classes in the IHAR dataset are displayed in Figure \ref{fig.IHARD}. To construct the IHAR dataset, we have incorporated different small-scale and medium-scale datasets, including Stanford40 \cite{yao2011human}, Pascal VOC 2012 Action\cite{everingham2015pascal}, BU101+ \cite{ashrafi2021action}, PPMI \cite{yao2010grouplet}, Sports \cite{gupta2009observing}, and ImSitu \cite{yatskar2016situation} datasets. Similar classes in the selected datasets are merged all together to form new classes in the IHAR dataset. For instance, the images relevant to the athletic movements in the selected datasets are integrated into the Sport class in the IHAR dataset. The number of classes chosen from each dataset is reported in Table \ref{ratio}.

\section{Experiments}\label{sec.6}

In this section, we initially describe the datasets, which are employed in our evaluations. Then, the implementation details and the exhaustive evaluation are addressed to prove the effectiveness of the proposed method. More specifically, subsection \ref{Experiments.Datasets} introduces different well-organized balanced datasets, which are employed in our assessments along with the IHAR dataset. Subsection \ref{Experiments.Implementation} provides the implementation details of the proposed action recognition approach. Subsection \ref{Experiments.Ablation} investigates the contribution of different proposed components in our action recognition approach. The performance of the proposed SCLAR on ablated versions of our method is addressed as well. Finally, quantitative and qualitative experiments are carried out in subsection \ref{Experiments.Comparisons}, and the comparison with the state-of-the-art studies is also provided in this section.

\begin{table}[!t]
\center
\caption{The number of different classes in the selected datasets, which has been utilized for the IHAR dataset construction.}
\resizebox{0.35\textwidth}{!}{%
\begin{tabular}{ccc}
\toprule
\toprule

&Dataset&     Utilized classes  \\ \midrule

&Stanford40 \cite{yao2011human}&All classes\\
 \addlinespace[1mm]

&Pascal VOC Action \cite{everingham2015pascal}&All classes\\
 \addlinespace[1mm]

&BU101+ \cite{ashrafi2021action} &20 classes\footnotemark\\
 \addlinespace[1mm]
&PPMI \cite{yao2010grouplet}&All classes\\
 \addlinespace[1mm]
&Sports \cite{gupta2009observing}&All classes\\
 \addlinespace[1mm]
&ImSitu \cite{yatskar2016situation}&23 classes\footnotemark\\
 \addlinespace[1mm]
\bottomrule
\bottomrule
\end{tabular}%
}

\label{ratio}%
\end{table}%

\footnotetext[1]{The classes include baseball pitch, cutting in kitchen, frisbee, playing piano, playing violin, rafting, rowing, soccer penalty, taiChi, writing on board, basketball, biking, brushing teeth, playing cello, playing flute, playing sitar, walking with dog, horse riding, playing daf, playing guitar, and playing tabla.}
\footnotetext[2]{The classes include sleeping, gardening, reading, praying, eating, shaving, driving, cooking, protesting, crowning, eating, bowing, drinking, sewing, rowing a boat, writing on a book, applauding, taking photos, phoning, eating, gardening, sport, and fishing.}

\subsection{Datasets}\label{Experiments.Datasets}
Apart from the introduced IHAR dataset, three other datasets (Stanford 40 \cite{yao2011human}, Pascal VOC 2012 Action \cite{everingham2015pascal}, BU101+ \cite{ashrafi2021action}) are also employed in our exhaustive evaluations. In what follows, we briefly review the datasets separately. In addition, PPMI, Sports, and ImSitu datasets are explained as follows:

\paragraph{Stanford 40} This dataset \cite{yao2011human} includes 9532 images in total corresponding to 40 classes of actions, which covers various real-world action classes such as riding a horse, waving
hands, running, and shooting an arrow. Eleven action classes are relevant to human body motion, and the others are based on non-body human motion. In each action class, 100 images are selected for the training phase, and the others are utilized for the testing phase.

\paragraph{Pascal VOC 2012 Action} Pascal VOC \cite{everingham2015pascal} composes of 4588 images in 10 classes, which is divided into 2296 training images, 2292 test images, and 4569 validation images. The images in Pascal VOC 2012 Action cover a large number of activities, including working with a computer, running, riding a horse, taking photos, and playing an instrument. The number of images in each action class for training and testing operations ranges from 140 to 220.

\paragraph{BU101+} This dataset \cite{ashrafi2021action} is the improved version of the BU101 dataset. To be more specific, it improves the cardinality of more than 90 classes of the original BU101 dataset. In addition, different challenges like dealing with background clutter and confusing objects are also taken into account in the BU101+  dataset. This dataset is constructed from 10100 images with 101 different action classes.

\paragraph{PPMI} The PPMI dataset \cite{yao2010grouplet} is constructed from different images of humans that interact with various musical instruments such as violin, clarinet, guitar, harp, recorder, and flute.

\paragraph{Sports} The Sports dataset \cite{gupta2009observing} contains a limited number of activity classes, including tennis forehand, croquet, tennis serve, and cricket batting. The classes are divided into 30 and 20 images for training and test phases. All images in the test and training sets are cropped and centered such that the persons of interest occupy a large proportion of the input images.

\paragraph{ImSitu} This dataset \cite{yatskar2016situation} includes 125000 images with 200000 distinctive situations. Each situation corresponds to one of 500 activities, 11000 objects, and 1700 roles. There are lots of images in this dataset that are suitable for human action recognition. We extracted several classes in this dataset to build the IHAR dataset.

\subsection{Implementation details}\label{Experiments.Implementation}
The proposed SCLAR contains two-phase training. In the first stage, the FGC-B and FAM components are required to be trained separately. The architecture of the FGC-B and FAM components follows the EfficientNET-B0 network, which has been pre-trained on the ImageNet \cite{deng2009imagenet}. Each branch of FGC-B is then trained for 100 epochs on an individual super-class which has been previously formed. In addition, the FAM component is trained for 100 epochs by means of all super-classes to distinguish different super-classes from each other at the coarse-grained phase. To train FGC-B, and FAM, we use Stochastic Gradient Descent (SGD) optimization with a momentum of 0.9, a weight decay of 20, and a batch size of 40. To train the FGC-B and FAM components, we set the learning rate to 0.5 and 0.1 in the Pascal VOC 2012 Action dataset and the remaining datasets, respectively. In the second phase, different branches of FGC-B are weighted by CGC and fed to the FGC-F component. Then, the FGC-F are trained from scratch same as FGC-B to determine the final output for action recognition. To train FGC-B, FAM and FGC-F components from scratch, the input images are first resized to $224\times 224$ pixels. Then, the overall architecture is fine-tuned with the input images with $448\times 448$ pixels. Also, two types of data augmentations are employed, including random horizontal flipping and random cropping. The optimum number of super-classes $M$  is set to 4 in our experiments. All the experiments are conducted on a single NVIDIA Geforce GTX 2080 TI GPU with 11 GB memory and the PyTorch toolbox.

\subsection{Ablation study}\label{Experiments.Ablation}
In this section, we conduct extensive ablation experiments to demonstrate the effectiveness of the key modules proposed in our SCLAR. All experiments in this part have been carried out on the Stanford40 dataset, which is the most widely benchmarked dataset in still image action recognition.

\begin{figure}[!t]
  \centering
  \captionsetup[subfigure]{font=footnotesize,justification=centerlast}
      \begin{subfigure}[c]{0.45\textwidth}
    \includegraphics[width=\textwidth]{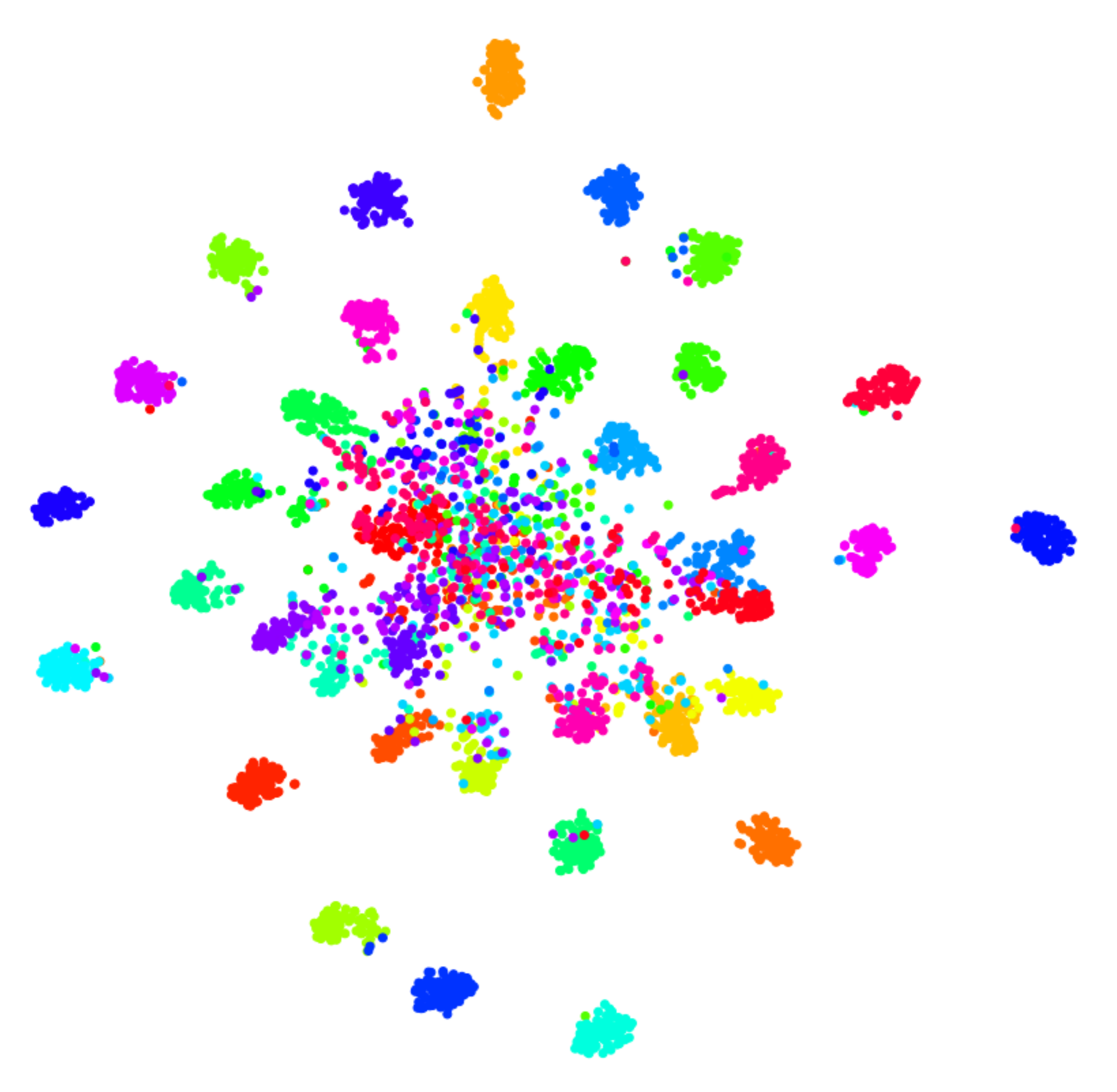}
    \caption[Short caption]{tSNE visualisation for the baseline model on the entire dataset.}
  \end{subfigure}
  \vfil
   \vspace{15pt}
  \begin{subfigure}[c]{0.35\textwidth}
    \includegraphics[width=\textwidth]{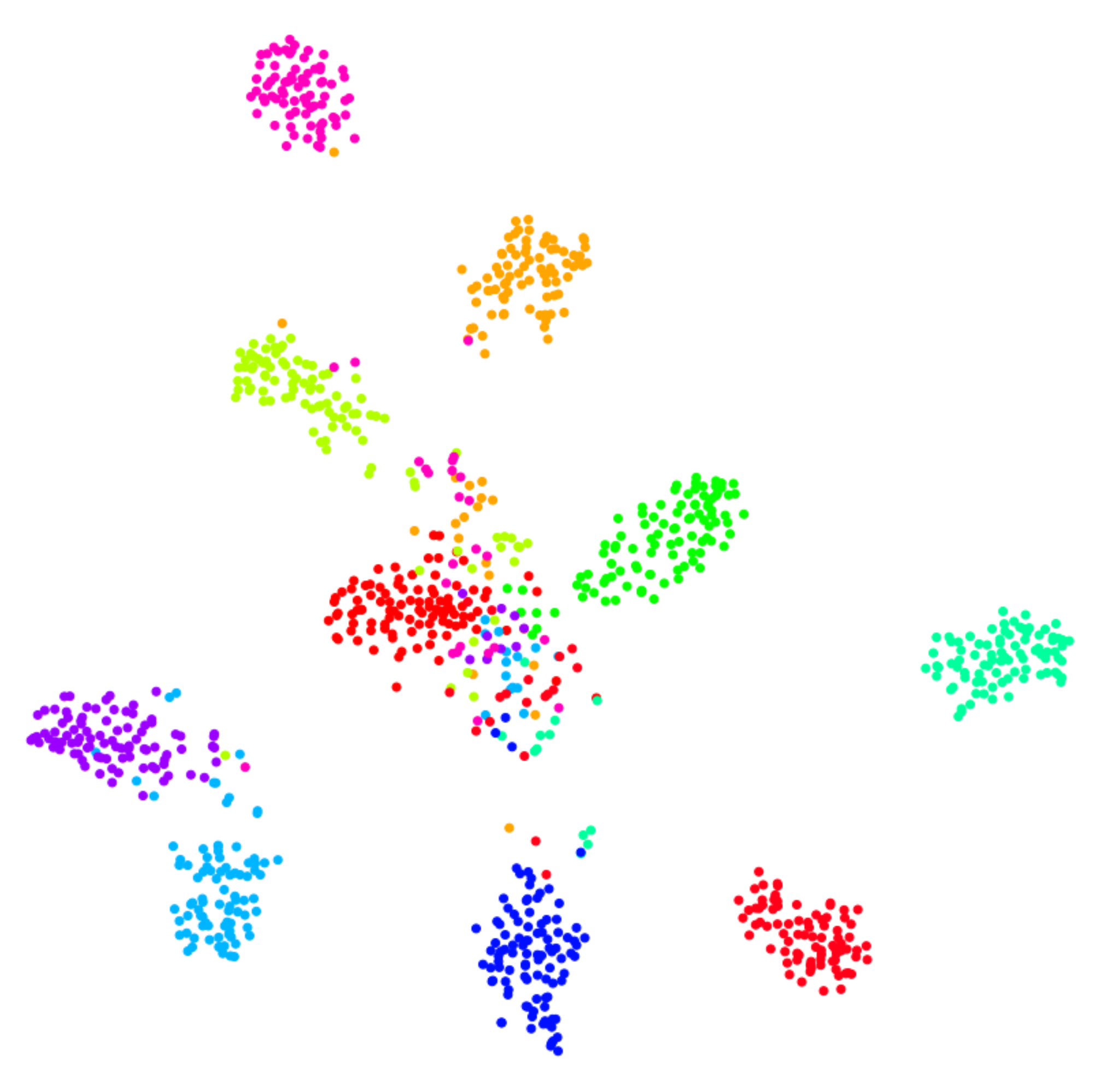}
    \caption[Short caption]{tSNE visualisation for the first expert on the first super-class.}
  \end{subfigure}
  \hspace{10pt}
  \begin{subfigure}[c]{0.35\textwidth}
    \includegraphics[width=\textwidth]{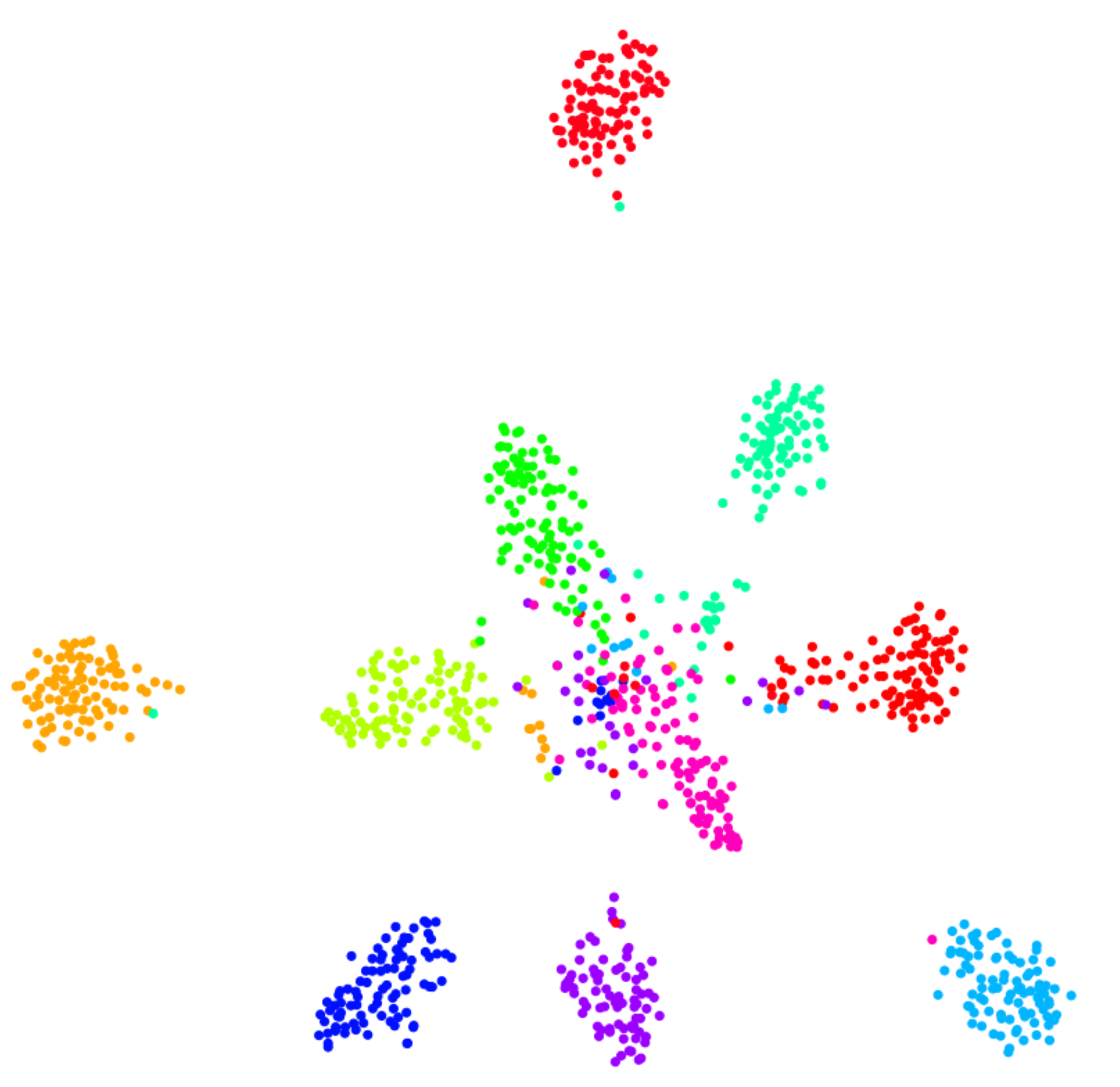}
    \caption[Short caption]{tSNE visualisation for the second expert on the second super-class.}
  \end{subfigure}
  \vfil
  \vspace{15pt}
    \begin{subfigure}[c]{0.35\textwidth}
    \includegraphics[width=\textwidth]{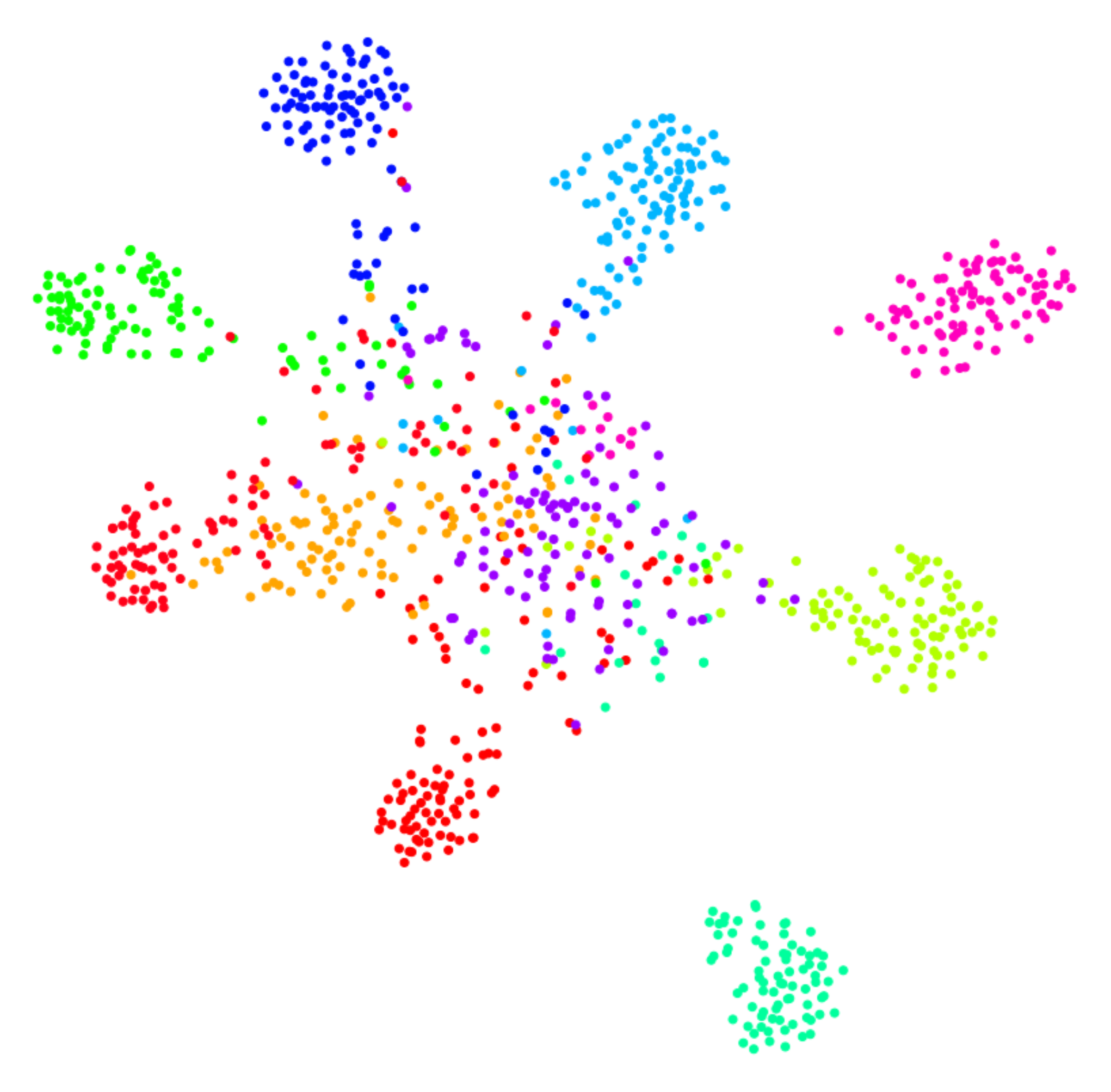}
    \caption[Short caption]{tSNE visualisation for the third expert on the third super-class.}
  \end{subfigure}
  \hspace{10pt}
  \begin{subfigure}[c]{0.35\textwidth}
    \includegraphics[width=\textwidth]{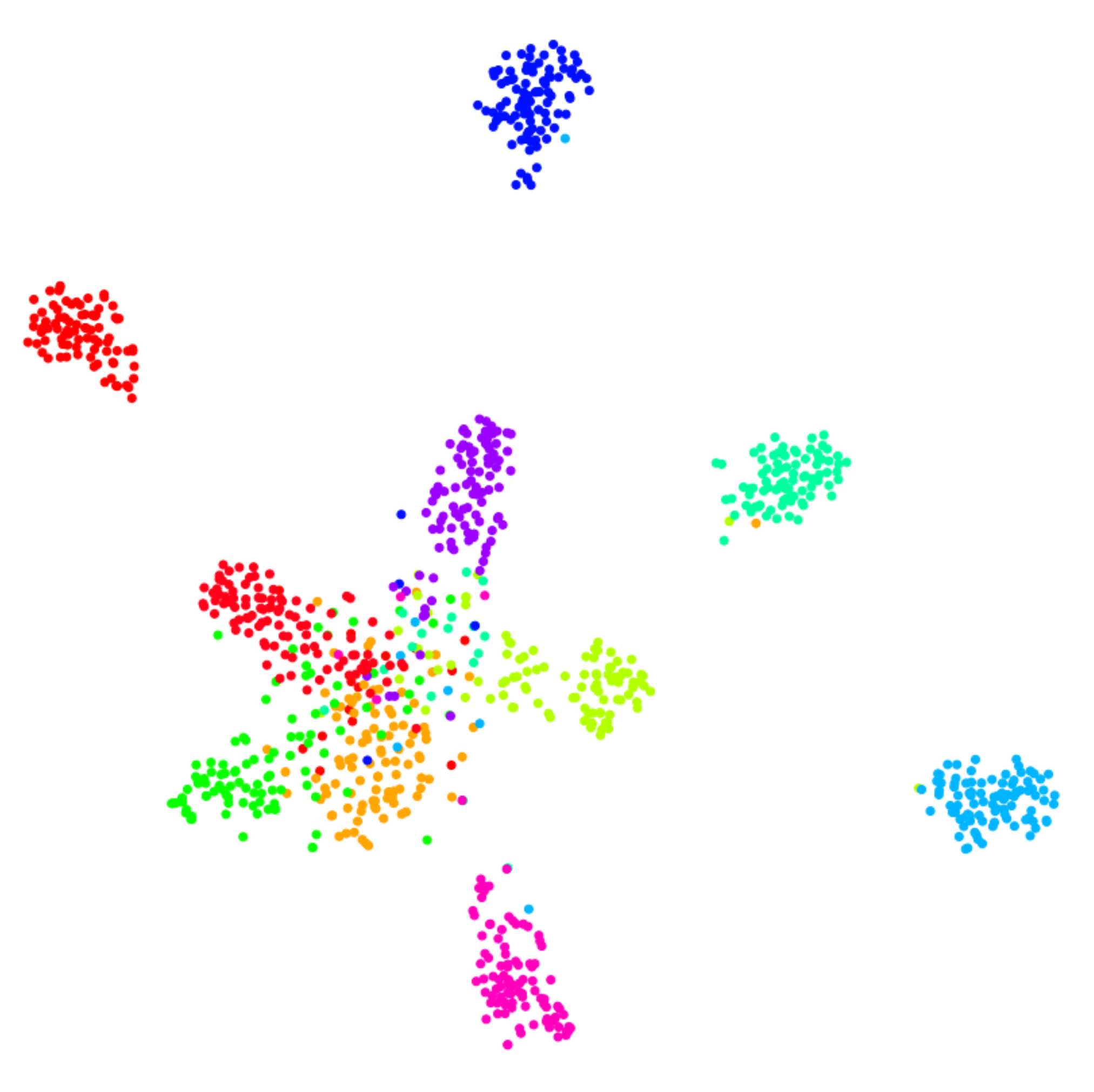}
    \caption[Short caption]{tSNE visualisation for the forth expert on the forth super-class.}
  \end{subfigure}

  \caption{tSNE visualisation for the features learned by the baseline model and different experts in the main architecture.}\label{tSNE}
\end{figure}

\begin{figure}[!t]
  \centering
  \includegraphics[scale=.35]{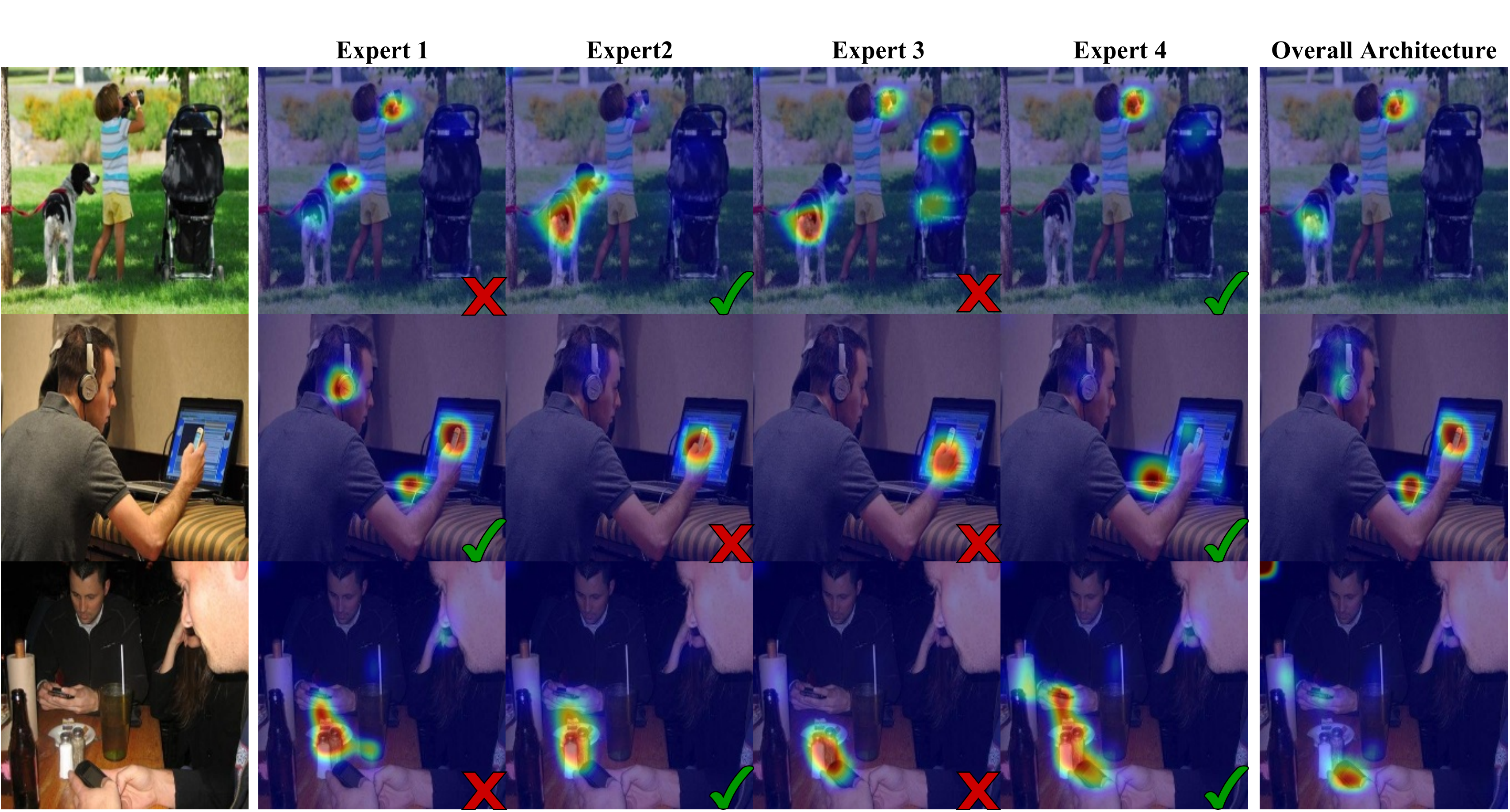}
  \caption{The heatmaps computed from different expert backbones and the main architecture with CGC.}\label{fig.heatmap_ab}
\end{figure}

\paragraph{Effect of GCS method for Super-Class Division} To investigate the impact of the generated super-classes on the GCS method, we set the number of super-classes to $M=4$ and assess the performance of each expert backbone in the FGC-B separately in comparison with the baseline model. Note that the baseline model is a single EfficientNet-B0 network which has been pre-trained over all classes. The results of this evaluation are reported in Table \ref{self_2}. The results demonstrate that each expert backbone yields better performance over the dedicated super-classes compared to the baseline model over all classes. This is largely ascribed to the fact that each expert backbone can extract more discriminative features from each super-class compared with the baseline model. In addition, to visually investigate this contribution, Figure \ref{tSNE} illustrates the tSNE visualization for different variations of expert numbers. It is observed that there exists low inter-class variation and no explicit boundaries for the baseline classification model within many classes; yet, the pre-trained expert backbones enjoy more discriminative representations and consequently can easily distinguish different action classes from each other. To explore the performance of different expert backbones for different inputs, we visualize the gradients of the top-class predictions in Figure \ref{fig.heatmap_ab} with respect to various input images by means of Smooth Grad-CAM++ \cite{omeiza2019smooth}. The green checkmarks denote the expert backbones which are selected by CGC. The red cross marks represent the other irrelevant expert backbones. As shown in Figure \ref{fig.heatmap_ab}, for the \textit{drinking} sample in the first row, the top selected expert ($\#4$) highlights the most informative image regions crucial to the proper classification. The other experts, on the other hand, concentrate on the areas uncorrelated with scenes and action-related objects; thus, the proposed CGC attempts to alleviate their impacts. In addition, in the third row, although the top expert ($\#2$) focuses on the bottles as well as the cell phone, the integration of all experts (the main proposed architecture) captures more discriminative areas with tighter bounds.

We also launch a study to benchmark the defectiveness of Equation \ref{eq.15} for finding the optimum configuration for all super-classes. In this study, we form the configuration of different super-classes randomly multiple times through which we assess the performance of action classification and report the average accuracy for all experiments. Then, we draw a comparison in Table \ref{random} between the results obtained with randomly configured super-classes and the proposed GCS method. Table \ref{random} validates that the proposed GCS method could find the optimum super-classes arrangement, thereby boosting the action recognition performance.

\begin{table}[!t]
\center
\caption{The impact of GCS method for finding the optimum configuration for different super-classes in the Stanford40 dataset.}
\resizebox{0.4\textwidth}{!}{%
\begin{tabular}{ccc}
\toprule
\toprule

& Super-Class Division Method &      mAP (\%) \\ \midrule

&Random Division &  88.7\\
&GCS Division &  92.86\\

\bottomrule
\bottomrule
\end{tabular}\label{random}}

\end{table}%

\paragraph{Effect of CGC Method} To demonstrate the impact of CGC in our main architecture, we deactivate this component and concatenate the output of expert backbones for the final classification. In this experiment, in the absence of CGC, different number of expert backbones is also ablated, as represented in Table \ref{self_3}. The results verify that CGC manages to assign the best expert to the relevant FGC in the fine classification phase. Interestingly, we observe that without CGC, multi-expert architecture lags behind the baseline model, and even the larger number of experts degrades the classification performance.

\begin{table}[!t]
\center
\caption{The impact of GCS method on different expert backbones for generating specialized super-classes in the Stanford40 dataset.}
\resizebox{0.4\textwidth}{!}{%
\begin{tabular}{cccc}
\toprule
\toprule

 &Model& Dataset&     mAP (\%) \\ \midrule

&Baseline&All classes&  83.5\\
 \addlinespace[1mm]
&Expert 1&Super-class 1 &93.5\\
 \addlinespace[1mm]
 &Expert 2&Super-class 2&93.9\\
 \addlinespace[1mm]
 &Expert 3&Super-class 3&94.2\\
 \addlinespace[1mm]
 &Expert 3&Super-class 4&93.2\\
 \addlinespace[1mm]
\bottomrule
\bottomrule
\end{tabular}\label{self_2}
}

\end{table}%

\begin{table}[!t]
\center
\caption{The impact of CGC method in the main architecture on action classification in the Stanford40 dataset.}
\resizebox{0.5\textwidth}{!}{%
\begin{tabular}{ccccc}
\toprule
\toprule

 &Model& Number of Experts&  CGC &   mAP (\%) \\ \midrule

&Baseline&\XSolidBrush &\XSolidBrush & 83.5\\
 \addlinespace[1mm]
&Multi-Expert&3&\XSolidBrush&82.4 \\
 \addlinespace[1mm]
 &Multi-Expert&4&\XSolidBrush&81.5 \\
 \addlinespace[1mm]
 &Multi-Expert&5&\XSolidBrush& 79.63\\
 \addlinespace[1mm]
 &Multi-Expert&4&\CheckmarkBold&92.86 \\
 \addlinespace[1mm]
\bottomrule
\bottomrule
\end{tabular}\label{self_3}
}

\end{table}%

\paragraph{Effect of M and S} In this experiment, the number of super-classes $M$ along with the threshold of the decision $S$ in CGM (Equation \ref{eq.2}) are investigated. Various experiments are performed with $M = 3, 4, 5$ and $S=1, 2, 3$. As reported in Table \ref{self_4}, $M=4$ and $S=2$ provide the best results in term of mAP metric. This study indicates that the larger values of $M$ and $S$ would not necessarily promote the performance of the proposed model.

\begin{table}[!t]
\center
\caption{The impact of $M$ (the number of super-classes) and $S$ (the threshold in Equation \ref{eq.2} in CGM) for action classification. The reported numbers in the table are the mAP result for action recognition in the Stanford40 dataset}
\resizebox{0.35\textwidth}{!}{%
\begin{tabular}{cccccc}
\toprule
\toprule

 && S = 1   & S = 2 & S = 3 &  \\ \midrule

&M = 3 &81.36 &89.6& 82.4 \\
 \addlinespace[1mm]
&M = 4 &82.4 &92.86 &85.6 \\
 \addlinespace[1mm]
 &M = 5 &81.95 & 90.4 &83.5\\
\bottomrule
\bottomrule
\end{tabular}\label{self_4}}

\end{table}

\subsection{Comparisons with state-of-the-art methods}\label{Experiments.Comparisons}
In this section, quantitative and qualitative experiments are performed to assess the contributions of SCLAR and prove its superiority compared with other studies. To this end, different benchmark datasets are adopted, including Stanford40, IHAR, BU101+, and Pascal VOC 2012 Action datasets.

\subsubsection{Stanford40}

\begin{table}[!t]
\center
\caption{Comprehensive results of the proposed method compared to the state-of-the-art action recognition works. The results are in terms of mean average precision (mAP) on the Stanford40 dataset. }

\resizebox{1\textwidth}{!}{%
\begin{tabular}{ccccccc}
\toprule
\toprule

&  Method & mAP (\%) & Bounding Box &Pose Estimator & Object Detector& Number of Parameters (M)\\ \midrule
 \addlinespace[1mm]

&Minimum Annotation \cite{ActionZhang} & 82.64&&& &24$>$\\
 \addlinespace[1mm]

 &Attention-Joints Graph \cite{ahmad2019action} & 84.6&\CheckmarkBold&& &$24>$\\
 \addlinespace[1mm]

&Deep VLAD Spatial Pyramids \cite{YAN2017118}  & 88.5&&& \CheckmarkBold&$24>$\\
 \addlinespace[1mm]

 &Pose-Guided  \cite{mi2021pose} & 89.53 &&\CheckmarkBold&\CheckmarkBold &$43>$\\
 \addlinespace[1mm]

 &BSE \cite{LI2020107341}   & 90.4&&\CheckmarkBold& &$43>$\\
 \addlinespace[1mm]

 &Multi-branch Attention \cite{8214269}    & 90.7&&\CheckmarkBold& &$30>$\\
 \addlinespace[1mm]

&Ensemble  \cite{mohammadi2019ensembles}  & 90.83&&& &$80>$\\
 \addlinespace[1mm]

&VIT \cite{dosovitskiy2020image}&  91.24& && &$85>$\\
 \addlinespace[1mm]

&Semantic Body Part \cite{zhao2017single}  & 91.2&&\CheckmarkBold&&$34>$\\
 \addlinespace[1mm]
&Loss Guided  \cite{liu2018loss}  & 91.2&&\CheckmarkBold& &$51>$ \\
 \addlinespace[1mm]

&Body-Part-aware \cite{bhandari2020body}  & 91.91&&\CheckmarkBold& &$64>$ \\
 \addlinespace[1mm]

&Multi-Attention Guided \cite{ashrafi2021action}  & 92.45 & \CheckmarkBold&& &$44>$\\
 \addlinespace[1mm]

&SCLAR& 92.86&&& &$20$\\

\bottomrule
\bottomrule
\end{tabular}\label{results.stanford}}
\end{table}
\begin{figure}[!t]
  \centering
  \includegraphics[scale=.37]{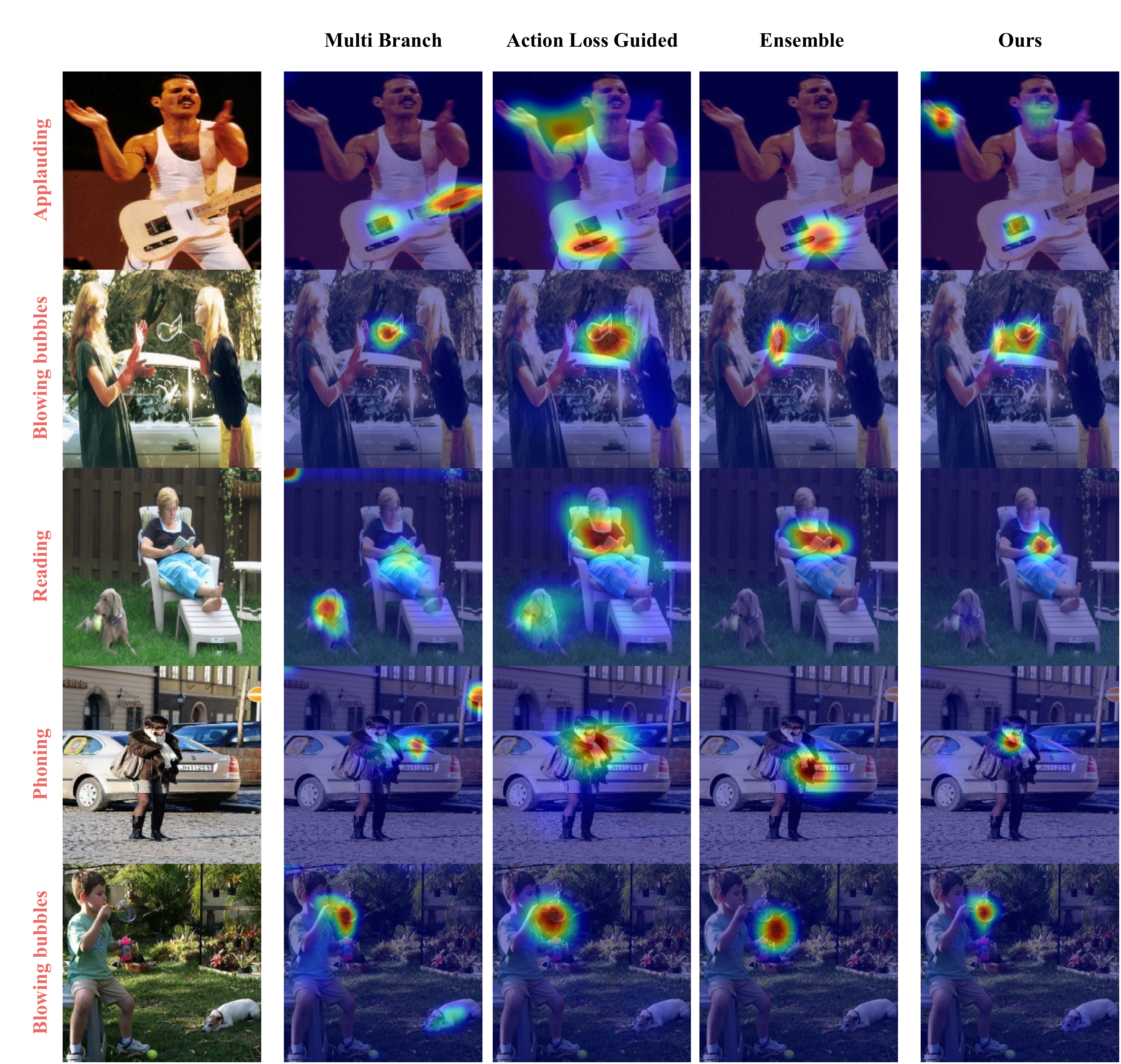}
  \caption{The heatmaps computed from different state-of-the-art studies and the proposed SCLAR.}\label{fig.heatmap_res}
\end{figure}

Table \ref{results.stanford} reports the mean average precision results of our action recognition method compared to its competitors on the Stanford40 dataset. To make an apple-to-apple comparison in still image action recognition, we also report different types of auxiliary components utilized in different studies in Table \ref{results.stanford}. The state-of-the-art studies in still image action recognition mainly utilize auxiliary components such as a pose estimator or object detector to provide the action classifiers with detailed information about different agents in the environment. Though the auxiliary components promote an accurate action recognition, they detrimentally make their approaches restricted to domain-specific data and also induce a heavy computation burden. As reported in Table \ref{results.stanford}, SCLAR yields the best performance in terms of accuracy and efficiency in comparison with state-of-the-art works. More specifically, as a runner-up, SCLAR yields gains of 0.41\% and 0.66\% on mAP compared to the second  \cite{ashrafi2021action} and third \cite{bhandari2020body} best methods. While the obtained gains in respect to the second-best method \cite{ashrafi2021action} is not significantly pronounced, the efficiency gain is dramatically considerable. Thus, SCLAR demonstrates a better accuracy/speed trade-off generally. Moreover, SCLAR relaxes the requirement of human bounding boxes in still images. Yet, \cite{ashrafi2021action} and \cite{bhandari2020body} require additional supervision, such as human bounding boxes or predefined body parts, which confine their practical applications to specific domains with available bounding boxes. Besides, it is proven that SCLAR outperforms other studies \cite{liu2018loss,8214269,LI2020107341,mi2021pose} to a great degree. The primary reason behind such noticeable superiority is that SCLAR could discriminate subtle visual inter-class variations from intra-class ones for action recognition task thanks to the CGC and FGC components. Note that although other studies \cite{liu2018loss,8214269,LI2020107341,mi2021pose,YAN2017118,mi2021pose} enjoy an additional object detector or pose estimator to localize body keypoints and filter out noisy context, they considerably lag behind SCLAR in both accuracy and efficiency metrics.

To qualitatively assess the proposed method compared to its counterpart, the gradients of top-class predictions are illustrated for different samples in Figure \ref{fig.heatmap_res}. It can be observed that the proposed method extracts tighter and more relevant regions of human agents in comparison with other studies. For instance, for \textsl{phoning}, and \textsl{reading} samples, the proposed SCLAR captures the action-related interactive objects (book and cell phone) accurately and attends less to the background regions. Moreover, not all areas in an image are helpful for action recognition. Irrelevant details captured by feature extraction phase in Loss Guided \cite{liu2018loss} and Multi-Branch \cite{8214269} hurt the performance of action recognition in their works. In challenging samples such as \textit{applauding}, SCLAR pays less attention to distractor objects. We highlight that distractor objects belong to irrelevant action classes in input images, thereby inducing misclassification.

\begin{figure}[!t]
  \centering
  \includegraphics[scale=.47]{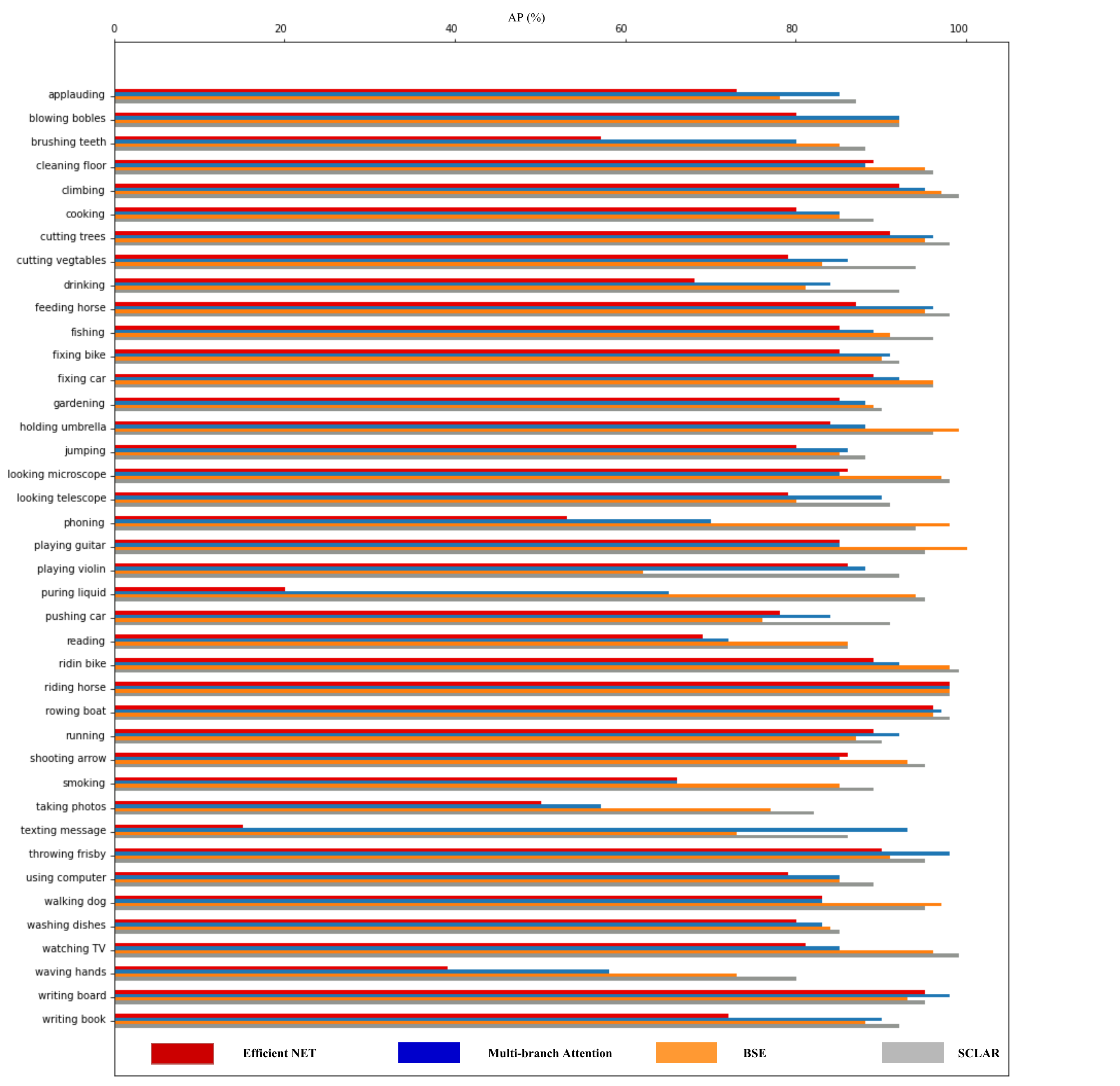}
  \caption{Results of the proposed SCLAR in comparison with the baseline EfficientNet and Multi-branch Attention \cite{8214269} on different classes of the Stanford40 dataset.}\label{fig.ap_classes}
\end{figure}

In addition, to evaluate the impact of SCLAR on different challenging classes with low inter-class variation, we compute average precision for all classes in the Stanford40 dataset. Figure \ref{fig.ap_classes} depicts the average precision (AP) scores of SCLAR along with single EffitientNet and the state-of-the-art study over all classes in the Stanford40 dataset. We analyze the least AP scores for different classes as well as their second-dependency classes. This analysis reveals that the proposed SCLAR gains substantial improvements over the baseline EffitientNet on the most challenging classes, which contain low inter-class variation with their second-class dependency classes. For instance, \textit{purring liquid}, \textit{phoning}, \textit{texting message}, \textit{taking photo}, and \textit{waving hand} classes possess low inter-class variations with their second dependency classes, which include \textit{washing dishes}, \textit{taking photo}, \textit{smoking}, \textit{phoning}, and \textit{smoking}, respectively. The proposed SCLAR outperforms the baseline EffitientNet (and also the state-of-the-art study) over these classes more significantly. In short, taking these results into account, we can substantiate the second contribution of this study which claims that SCLAR surpasses its competitors in HAR while requiring much less computational cost.

\subsubsection{PASCAL VOC 2012 Action}
Table \ref{results.PASCAL} presents the performance of SCLAR on the PASCAL VOC 2012 Action dataset. As reported in Table \ref{results.PASCAL}, the superiority of SCLAR is
kept with much less computational cost even compared to various state-of-the-art studies \cite{ashrafi2021action,LI2020107341}, which leverage from auxiliary supervision such as annotated bounding boxes or auxiliary modules such as pose estimator. To be more specific, the Multi-Attention Guided \cite{ashrafi2021action} approach adopts a teacher-student knowledge distillation which requires annotated bounding boxes and huge computation cost. In addition, the BSE \cite{LI2020107341} extracts body structure cues such as structural body parts and limb angle descriptors from local and global perspectives by means of a complicated three-branch classifier.

\begin{table}[!t]
\center
\caption{Comprehensive results of the proposed method compared to the state-of-the-art action recognition works. The results are in terms of mean average precision (mAP) on the PASCAL VOC 2012 Action dataset. }

\resizebox{1\textwidth}{!}{%
\begin{tabular}{ccccccc}
\toprule
\toprule

&  Method & mAP (\%) & Bounding Box &Pose Estimator &  Number of Parameters (M)\\ \midrule
 \addlinespace[1mm]

&Minimum Annotation \cite{ActionZhang} & 83.23&& &24$>$\\
 \addlinespace[1mm]

 &SAAM-Nets \cite{ZHENG2020383} & 84.8&& &$25>$\\
 \addlinespace[1mm]

 &Multi-branch Attention \cite{8214269}    & 87.1&&\CheckmarkBold& $30>$\\
 \addlinespace[1mm]

 &R*CNN \cite {gkioxari2015contextual} & 90.4&&& $24>$\\
 \addlinespace[1mm]

&Multi-Attention Guided \cite{ashrafi2021action}  & 91.51 & \CheckmarkBold&& $44>$\\
 \addlinespace[1mm]
 &BSE \cite{LI2020107341}   & 91.8&&\CheckmarkBold& $43>$\\
 \addlinespace[1mm]

&SCLAR& 92.46&&& 20\\

\bottomrule
\bottomrule
\end{tabular}\label{results.PASCAL}}

\end{table}%

\subsubsection{BU101+}
To gauge the generalization of SCLAR, we also evaluate SCLAR on the BU101+ dataset. This dataset offers new challenges such as background clutter and confusing human agents for the action recognition evaluation in compariosn with the standard Stanford40 dataset. Since BU101+ has been introduced recently, limited research works have been benchmarked on this dataset. As such, we suffice to pursue our comparisons the same as \cite{ashrafi2021action}. Based on Table \ref{results.BU101}, SCLAR obtains better results than Multi-Attention Guided research work \cite{ashrafi2021action} even when \cite{ashrafi2021action} accesses to annotated bounding boxes. Considering reported results in Table \ref{results.BU101}, we can demonstrate the effectiveness of the proposed GCS method in SCLAR to strengthen the inter-class variation.

\begin{table}[!t]
\center
\caption{Comprehensive results of the proposed method compared to the state-of-the-art action recognition works. The results are in terms of mean average precision (mAP) on the BU101+ dataset.}

\resizebox{0.8\textwidth}{!}{%
\begin{tabular}{ccccccc}
\toprule
\toprule

&  Method &   mAP (\%) &Number of Parameters (M)\\ \midrule
 \addlinespace[1mm]

&ResNET\_18  \cite{he2016deep}& 83.06& 11.5\\
 \addlinespace[1mm]

&ResNET\_34  \cite{he2016deep}& 87.07& 21.6\\
 \addlinespace[1mm]

 &ResNET\_50  \cite{he2016deep}& 87.44& 24\\
 \addlinespace[1mm]

&Multi-Attention Guided \cite{ashrafi2021action}  & 90.16 & $44>$\\
 \addlinespace[1mm]

&SCLAR& 92.27& 20\\

\bottomrule
\bottomrule
\end{tabular}\label{results.BU101}}

\end{table}%

\subsubsection{IHAR}

In the last part of our evaluations, we seek to study whether the proposed SCLAR can handle the data imbalance issue for action recognition with
long-tailed distribution in still images. Table \ref{results.IHAR} depicts our comparisons on the IHAR dataset. The proposed lightweight SCLAR achieves 92.27 mAP accuracy and remarkably outperforms the state-of-the-art models with a gain of 8.92\% compared to the second-best study \cite{mohammadi2019ensembles}. The reason behind such a pronounced gain is that the GCS would divide the input dataset into different super-classes with relatively balanced distribution such that different backbone experts can learn better from under-represented classes in super-classes. A comparison between our evaluations on the IHAR dataset (Table \ref{results.IHAR}) with the other balanced datasets in Table \ref{results.PASCAL} and  Table \ref{results.stanford} reveals that the superiority of SCLAR over the state-of-the-art studies such as Ensemble  \cite{mohammadi2019ensembles} is more noticeable on long-tailed distribution than conventional balance datasets. All in all, we can deduce that the contributions of GCS and CGC in dealing with data imbalance issue in still image action recognition with long-tailed distribution are fully corroborated by the reported results in Table \ref{results.BU101}.

\begin{table}[!t]
\center
\caption{Comprehensive results of the proposed method compared to the state-of-the-art action recognition works. The results are in terms of mean average precision (mAP) on the introduced IHAR dataset. }

\resizebox{0.8\textwidth}{!}{%
\begin{tabular}{ccccccc}
\toprule
\toprule

&  Method &      mAP (\%)&  Number of Parameters (M)\\ \midrule
 \addlinespace[1mm]

 &ResNET\_50   \cite{he2016deep}& 87.44&24\\
 \addlinespace[1mm]

&Multi-branch Attention \cite{8214269}  & 81.9  &$44>$\\
 \addlinespace[1mm]
 &R*CNN \cite {gkioxari2015contextual} & 83.02&$24>$\\
 \addlinespace[1mm]

&Ensemble  \cite{mohammadi2019ensembles}  & 83.35  &80\\
 \addlinespace[1mm]

&SCLAR& 92.27&20\\

\bottomrule
\bottomrule
\end{tabular}\label{results.IHAR}}

\end{table}%

\section{Conclusion}\label{sec.7}
In this paper, we propose a two-phase multi-expert architecture for still image action recognition, which contains fine-grained and coarse-grained phases. The fine-grained stage includes a combination of pre-trained experts, and the coarse-grained stage contains FAM to select the relevant fine-grained features from the fine-grained phase. To the best of our knowledge, the proposed two-phase multi-expert architecture is the first action recognition study that adaptively hedges features from pre-trained experts. Moreover, the GCS method is introduced to divide the input dataset into M super-classes for different experts in the fine-grained phase such that the inter-class variance increases. We also introduce a new still image action recognition dataset with long-tailed distribution to assess the robustness of our method against the data imbalance issue. Our experiments verify the effectiveness of different proposed components in our action recognition method and demonstrate its superiority compared with the state-of-the-art studies.

\appendix
\section{Example of GCS }\label{sec.app}

In this appendix, we present an example to illustrate GCS algorithm for partitioning various classes in our dataset. Let our dataset include different classes as $Classes = \{A,B,C,D,E,F,G,H,I,J,K,L,M,N,O\}$. Then, the second-order, third-order, and fourth-order dependency sets  for our dataset can be expressed as follows:

\begin{equation}\label{eq.x2}
D_{2} = \{B,C,E,C,F,J,F,F,F,K,J,K,L,O,N\},
\end{equation}

\begin{equation}\label{eq.x3}
D_{3} = \{J,N,G,B,A,H,L,N,D,M,A,E,A,K,J\},
\end{equation}

\begin{equation}\label{eq.x4}
D_{4} = \{M,C,E,A,L,J,B,N,G,K,D,N,E,O,E\},
\end{equation}

\noindent Figure \ref{fig.APP1} illustrates the constructed derivative graphs. To find the optimum configuration of different super-classes, the following steps are taken:
\begin{figure}[!t]
  \centering
  \includegraphics[scale=.9]{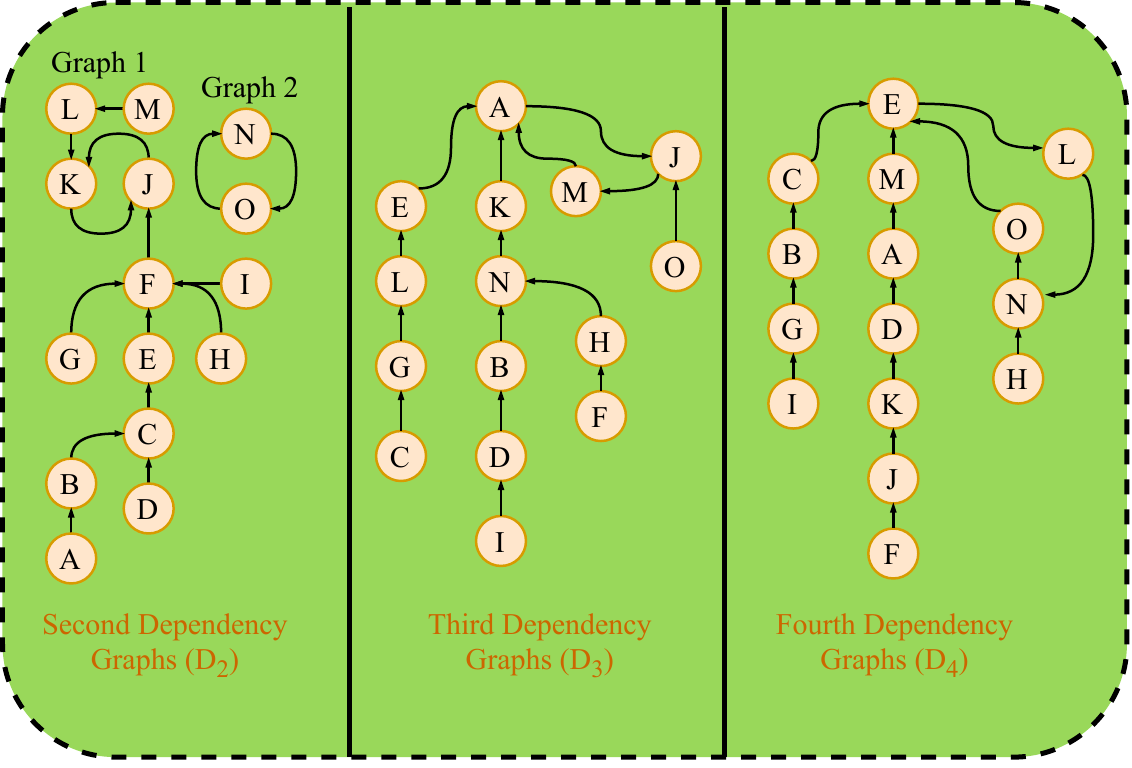}
  \caption{Different dependency graphs in the proposed GCS algorithm.}\label{fig.APP1}
\end{figure}

\begin{figure}[H]
  \centering
  \includegraphics[scale=1]{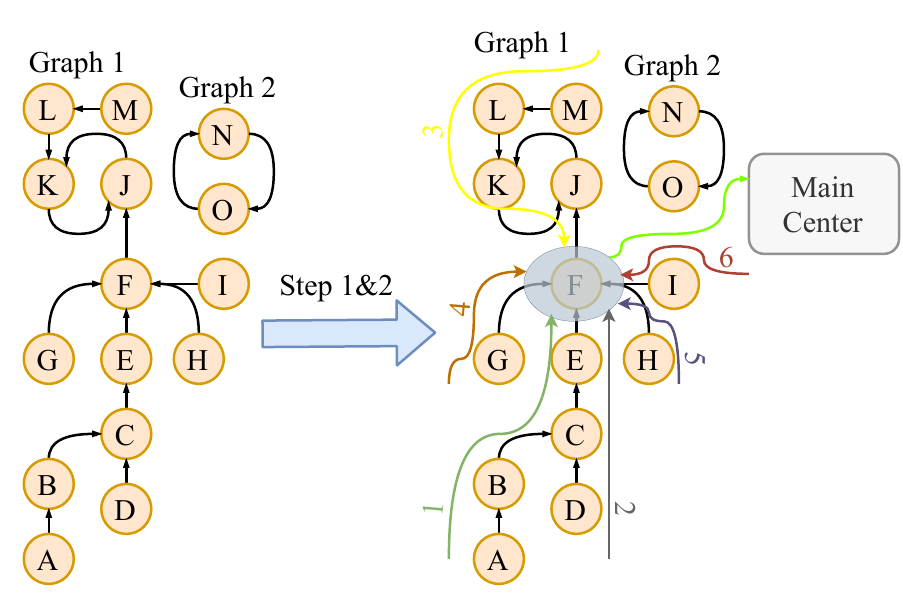}
  \caption{Illustration of the first and second steps in the GCS algorithm. The main center node has the largest number of connections. Node F is the main center node in this figure. In the second order graph ($D_{2}$), a set of connected nodes which end with the center node and begin from a boundary node is defined as a route. $ABCEF$, $GF$, and $MLKJ$ exemplify different routes which end with the main center node $F$.}\label{fig.APP2}
\end{figure}

\begin{itemize}

\item Finding the center node in second order graphs.

\item Numbering all routes in the second order graphs (Figure \ref{fig.APP2} shows the outcome of the first and second steps).

\item In the second order graphs ($D_{2}$), according to the route score, the connected nodes are placed into the sequential super-classes.

\item Optimizing the arrangement of classes in the super-classes according to $D_{2}$ such that the number of class pairs with $S_{1}$, and $S_{2}$ similarities would be minimized in each super-class.

\item Optimizing the arrangement of classes in the super-classes according to $D_{3}$ such that the number of class pairs with $S_{3}$ similarity would be minimized, and no new $S_{1}$, and $S_{2}$ similarities would be created.

\item Optimizing the arrangement of classes in the super-classes according to $D_{4}$ such that the number of class pairs with $S_{4}$ similarity would be minimized, and no new $S_{1}$, $S_{2}$, and $S_{3}$ similarities would be created.

\end{itemize}

\begin{figure}[H]
  \centering
  \includegraphics[scale=.5]{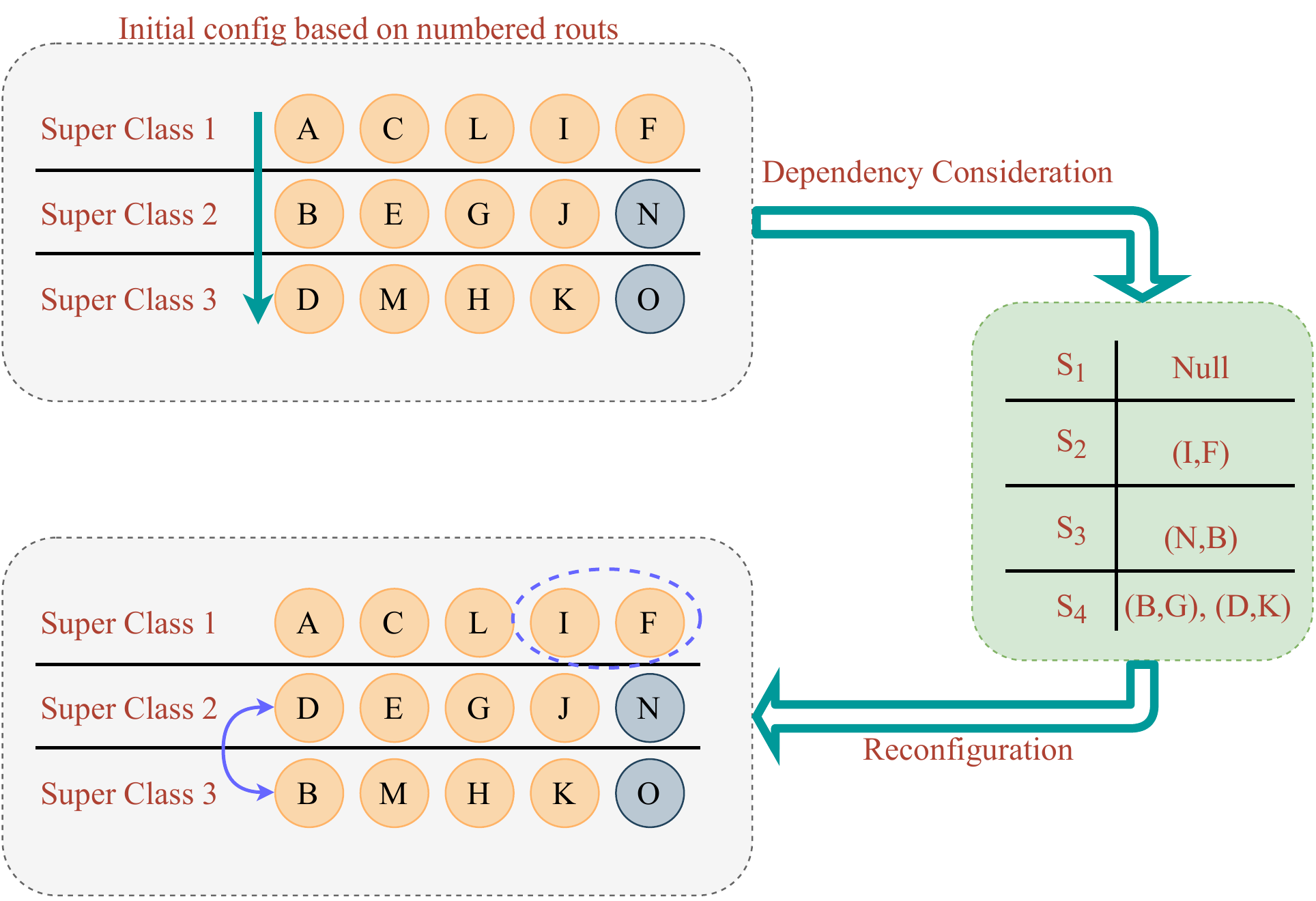}
  \caption{The arrangement of classes for the super-classes with $M=3$.}\label{fig.APP3}
\end{figure}

For this example, once there exist three super-classes, i.e., $M=3$, the arrangement of classes for the super-classes would be as Figure \ref{fig.APP3}. In the classes configuration, classes $I$ and $F$ contain $S_{2}$ similarity. Since rearranging each of $I$ and $F$ classes with other classes would lead to a new $S_{2}$ similarity with other classes, we would not change the configuration of these classes. However, pair classes such as $(N,B)$, $(D,K)$ and $(B,G)$ in the second and third super-classes contain $S_{3}$ and $S_{4}$ similarities, respectively. Rearranging these classes with others would properly eliminate their similarities.

\bibliography{bibfile}

\end{document}